\documentclass{article}

\usepackage{PRIMEarxiv}

\usepackage[utf8]{inputenc} 
\usepackage[T1]{fontenc}    
\usepackage{hyperref}       
\usepackage{url}            
\usepackage{booktabs}       
\usepackage{amsfonts}       
\usepackage{nicefrac}       
\usepackage{microtype}      
\usepackage{lipsum}
\usepackage{fancyhdr}       
\usepackage{graphicx}       
\usepackage{natbib}
\usepackage{caption}
\usepackage{subcaption}
\usepackage{multirow}
\usepackage{tabularx}
\usepackage{amsmath}
\graphicspath{{media/}}     

\DeclareMathOperator*{\argmax}{arg\,max}

\DeclareMathOperator{\softmax}{softmax}
\title{Unsupervised and Few-shot Parsing from Pretrained Language Models
}

\author{
  Zhiyuan Zeng \\
  Tianjin University \\
  Tianjin\\
  \texttt{zhiyuan\_zeng@tju.edu.cn} \\
  \And
   Deyi Xiong \\
  Tianjin University \\
  Tianjin \\
  \texttt{dyxiong@tju.edu.cn} \\
}

\begin{document}
\maketitle

\begin{abstract}
Pretrained language models are generally acknowledged to be able to encode syntax \citep{tenney-etal-2019-bert,jawahar-etal-2019-bert,hewitt-manning-2019-structural}. In this article, we propose UPOA, an Unsupervised constituent Parsing model that calculates an Out Association score solely based on the self-attention weight matrix learned in a pretrained language model as the syntactic distance for span segmentation. We further propose an enhanced version, UPIO, which exploits both inside association and outside association scores for estimating the likelihood of a span. Experiments with UPOA and UPIO disclose that the linear projection matrices for the query and key in the self-attention mechanism play an important role in parsing. We therefore extend the unsupervised models to few-shot parsing models (FPOA, FPIO) that use a few annotated trees to learn better linear projection matrices for parsing. Experiments on the Penn Treebank demonstrate that our unsupervised parsing model UPIO achieves results comparable to the state of the art on short sentences (length <= 10). Our few-shot parsing model FPIO trained with only 20 annotated trees outperforms a previous few-shot parsing method trained with 50 annotated trees. Experiments on cross-lingual parsing show that both unsupervised and few-shot parsing methods are better than previous methods on most languages of SPMRL \citep{spmrl}.
\end{abstract}

\keywords{Unsupervised constituent parsing \and Few-shot parsing \and Pretrained language model}

\section{Introduction}
Automatically parsing a sentence to unveil the latent syntactic structure of the sentence is a long-standing task in natural language processing. This is not only because language is inherently hierarchically structured \citep{Chomsky1957,de2011course}, but also due to the practical use of syntactic structures for a variety of downstream tasks, e.g., semantic parsing \citep{raymond2006discriminative,goli2017syntax}, machine translation \citep{liu-etal-2006-tree,mi-etal-2008-forest,chiang-2010-learning,li-etal-2017-modeling}, etc. Supervised syntactic parsing usually models grammars in a probabilistic \citep{collins2003head,klein2003accurate,petrov2007improved,Dyer_2016} or non-probabilistic way \citep{kitaev-klein-2018-constituency,stern-etal-2017-minimal}, which usually requires manually-annotated syntactic trees from a large treebank for training. However, building a treebank like PTB \citep{ptb} is expensive and time-consuming. Therefore such datasets are available only for a few languages and genres. 

Unsupervised constituent parsing,  which, related to grammar induction, learns underlying structures without using any annotated trees, hence becomes an alternative to supervised syntactic structure learning. With extensive interest in unsupervised grammar induction in recent years, this direction have proposed a variety of neural models for unsupervised parsing \citep{shen2018neural,shen2018ordered,drozdov-etal-2019-unsupervised,kim-etal-2019-unsupervised,kim-etal-2019-compound,Chen-2019-tree}, where latent structures are usually learned in a neural language model from unannotated sentences.  However, the unsupervised paradigm is still very challenging with a parsing accuracy that lags far behind the accuracy of its supervised counterpart. 

In this article, we propose a new framework for unsupervised constituent parsing. First, partially inspired by \citet{shen2018neural}, we define a new syntactic distance for Unsupervised constituent Parsing, which is calculated according to an Outside Association score solely based on self-attention weight matrix: \textbf{UPOA}. Previous findings from many probing studies suggest that pretrained language models \citep{devlin-etal-2019-bert,liu2019roberta,radford2019language} are able to learn and embed latent syntactic structures in their parameters in an implicit way \citep{tenney-etal-2019-bert,jawahar-etal-2019-bert}. Therefore, in UPOA, we exploit the self-attention weight matrix in BERT \citep{devlin-etal-2019-bert} to uncover latent syntactic structures of sentences.\footnote{We leave the exploitation of other pretranined language models for our future work.} Particularly, we suppose that the self-attention weights can be used for the approximation to the negative syntactic distance, since both of them measure the relatedness between two words. We estimate the syntactic distance between two adjacent words as the negative self-attention weight between two adjacent spans that consume the two words. With the estimated syntactic distance, UPOA splits a span at the split point with the largest syntactic distance. However the syntactic distance defined in UPOA only considers the association between adjacent spans (outside association), ignoring the association among words inside a span (inside association). Therefore, we further propose an enhanced Unsupervised Parsing model \textbf{UPIO}, which splits a span according to the strength of both Inside and Outside association. The inside association is also estimated with the self-attention weights. With the estimated inside association and outside association, we can build a syntactic tree for a sentence with a greedy algorithm or a chart parsing algorithm.

Second, we extend the UPIO to \textbf{FPIO}, a few-shot version of UPIO that can learn substantially better syntactic structures to narrow the performance gap between unsupervised and supervised parsing with just a few annotated trees. The fundamental idea behind FPIO is based on our finding with UPIO that the two linear projection matrices used by the query and key in the self-attention mechanism have a great impact on the parsing accuracy of the FPIO. We therefore freeze other parameters in BERT and propose a method to retrain (i.e., fine-tune) the two projection matrices on a few annotated trees. Similarly, we extend the UPOA to FPOA, a few-shot version of UPOA. The difference between FPOA and FPIO is that FPOA splits a span according to the syntactic distance (outside association), while FPIO exploits not only the outside association among words inside a span and words outside the span, but also the inside association among words within a span.

We carry out the unsupervised and few-shot parsing experiments on the Penn Treebank (PTB) \citep{ptb} and SPMRL \citep{spmrl} dataset. In addition to the monolingual unsupervised/few-shot parsing on PTB, we also conduct cross-lingual unsupervised/few-shot parsing on SPMRL, where the unsupervised or few-shot parsing model tuned/trained on PTB are evaluated on languages in SPMRL. Our contributions can be summarized as follows:
\begin{itemize}
\item We propose an unsupervised constituent parsing model UPOA that calculates an out association score for span segmentation solely based on the self-attention weight matrix in a pretrained language model, and further propose an enhanced model, UPIO, which exploits both inside and outside association scores for estimating the likelihood of a span.
\item We further extend our unsupervised parsing models UPOA and UPIO to few-shot learning methods FPOA and FPIO, which, trained on just a few annotated examples, substantially outperforms the few-shot and supervised parsing methods on the Penn Treebank and most languages of SPMRL.
\item We conduct experiments and in-depth analyses on the proposed unsupervised and few-shot parsing models, which not only demonstrate the effectiveness of both models, but also provide interesting findings, e.g., those with few-shot parsing on different layers and self-attention heads.
\end{itemize}

\section{Related Work}
Our work is related to : 1) unsupervised constituent parsing 2) particularly recent works on syntactic probing into pretrained language models and extracting latent trees from these pretrained models.

\paragraph{Unsupervised Constituent Parsing}  Unsupervised constituent parsing is a challenging task that induces hierarchical syntactic structures from unlabeled data and learns a probabilistic grammar generalizable to unseen sentences. In early work, a variety of statistical models have been proposed, such as generative models \citep{klein-manning-2002-generative,klein-manning-2004-corpus},  non-parametric models \citep{bod-2006-subtrees,seginer-2007-fast}, just to name a few.

Recent years have witnessed a resurgence of interest in unsupervised parsing with a methodological shift from traditional statistical models to neural approaches. These models are usually trained with tasks to optimize an unsupervised objective on a large amount of raw unlabeled data. \citet{shen2018neural} propose Parsing-Reading-Predict Networks (PRPN) to learn a better language model by leveraging structure information. \citet{shen2018ordered} explore Ordered Neurons trained on language models to learn latent tree structures, which introduces an inductive bias to the LSTM \citep{lstm} (ON-LSTM). 
\citet{jin-etal-2019-unsupervised} propose a neural PCFG inducer that employs context embeddings in a normalizing flow model to extend PCFG induction to use semantic and morphological information.
\citet{drozdov-etal-2019-unsupervised} present recursive auto-encoders with the inside-outside algorithm (DIORA). \citet{kim-etal-2019-compound} propose a compound probability context-free grammar to neural unsupervised parsing (C-PCFG). \citet{kim-etal-2019-unsupervised} propose unsupervised recurrent neural network grammars (URNNG) with amortized variational inference on a structured inference network. \citet{Chen-2019-tree} incorporate a ``Constituent Prior'' into the self-attention mechanism of Transformer which attempts to force words to attend other words in the same constituent (Tree-Transformer). 
\citet{jin-schuler-2020-grounded} and \citet{zhang-etal-2021-video} extract rich features from images and videos for grammar induction. \citet{li-etal-2020-empirical} have studied and compared some latest unsupervised parsing methods, and found that both the experiment setting and the evaluation metrics have a great impact on the performance of these methods. \citet{shi2020role} have found that most unsupervised parsing methods are highly relied on the hyper-parameters tuned on the validation set, which inspires us to carry out the cross-lingual experiments in Section \ref{section:Cross-lingual parsing Results}.

\paragraph{Parsing with Pretrained Language Models} 
Many probing studies \citep{tenney-etal-2019-bert,jawahar-etal-2019-bert} explore hidden representations learnt in pretrained language models for syntactic tasks, e.g., constituent labeling, which have found that the intermediate layers of pretrained language models can capture rich syntactic information. \citet{bert_look_at} exploit the attention matrix of BERT for predicting the dependencies and coreference of words, and demonstrate that substantial syntactic information is captured in BERT’s attention matrix. These findings trigger research interest of using pretrained language models for parsing. \citet{parsing_as_pretrain} train a supervised parser based on a pretrained language model, which does not need top-down or chart-based decoding, but still achieves good results. \citet{hewitt-manning-2019-structural} train a probe on BERT to predict the distance of words in dependency trees, and use the predicted distance for dependency parsing. \citet{htut2019attention} employ the maximum-spanning-tree algorithm on the self-attention weight matrix of pretrained language models to produce dependency trees. However, they find it is difficult to produce valid trees. \citet{marecek-rosa-2018-extracting} introduce a simple algorithm to generate constituency trees from the self-attention weight matrix of Transformer. They have only manually compared their results of 42 sentences to ground-truth trees in PTB and haven't report any experiment results. \citet{Kim2020Are} study the combination of different distance measure functions (e.g., {\em cosine, L1, L2}) and types of word representations from pretrained language models on grammar induction. Since their key interest is to investigate whether pretrained language models exhibit syntactic structures and to provide a baseline for unsupervised grammar induction from these pretrained language models, their results are not comparable to state-of-the-art unsupervised parsing models. \citet{heads_up} propose a fully unsupervised constituent parser, which ranks attention heads of pretrained language model, and create an ensemble of them for parsing. \citet{kimchart} extend the model proposed in \citep{Kim2020Are} to a CKY parser, and then evaluate their unsupervised parser on SPMRL \citep{spmrl} dataset in a cross-lingual setting.

In this article, we propose a parsing framework based on the self-attention mechanism of pretrained language models to extract latent constituency structures of sentences, which includes the unsupervised UPOA, UPIO and few-shot FPOA, FPIO. The parsing algorithm  used in this framework is similar to some previous supervised parsing algorithm \citep{stern-etal-2017-minimal,kitaev-klein-2018-constituency}. However, significantly different from them, our span scores are computed from the attention weight matrix instead of the hidden representations. 
\citet{Kim2020Are} also exploit the pretrained language model to estimate the syntactic distance for unsupervised parsing, which is similar to our UPOA. However, their syntactic distance is estimated with the hidden representations of two adjacent words, while the syntactic distance in UPOA is estimated with the average attention weights of two adjacent spans, which reduces the estimation bias of the syntactic distance. \citet{marecek-rosa-2018-extracting} also explore the self-attention mechanism of Transformer for tree extraction. However, unlike our UPIO, they only consider the association of words inside a constituent, ignoring the distance to words outside the constituent. We validate the importance of this distance information to parsing in our ablation study in Section \ref{Ablation-study}.

\section{Unsupervised Parsing from Pretrained Self-Attention}
In this section, we first elaborate our proposed methods to compute split scores based on the attention weights of self-attention mechanism from a pretrained language model and then introduce two parsing algorithms with the split scores, namely greedy parsing and chart parsing.

\subsection{Estimating Split Scores based on Syntactic Distance}\label{section:Split Scores based on syntactic distance}
\begin{figure}[t]
\centering
   \begin{subfigure}[b]{0.4\linewidth}	
   \centering
	\includegraphics[width=\linewidth]{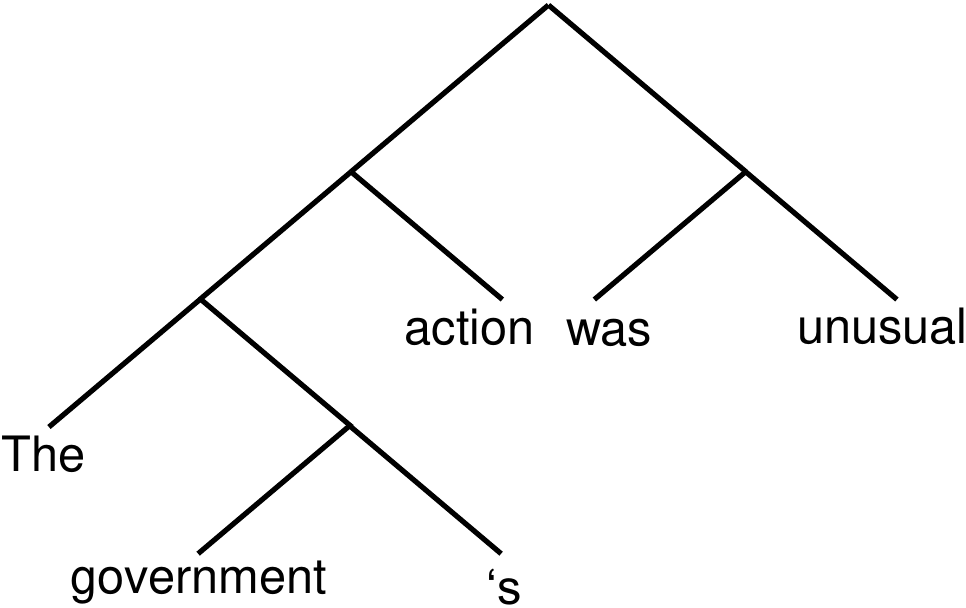}
	\setlength{\abovecaptionskip}{15pt plus 3pt minus 2pt}
	\caption{}
    \label{fig:tree-matrix-a}
	\end{subfigure}
    \quad\quad
   \begin{subfigure}[b]{0.4\linewidth}	
     \centering
	\includegraphics[width=\linewidth]{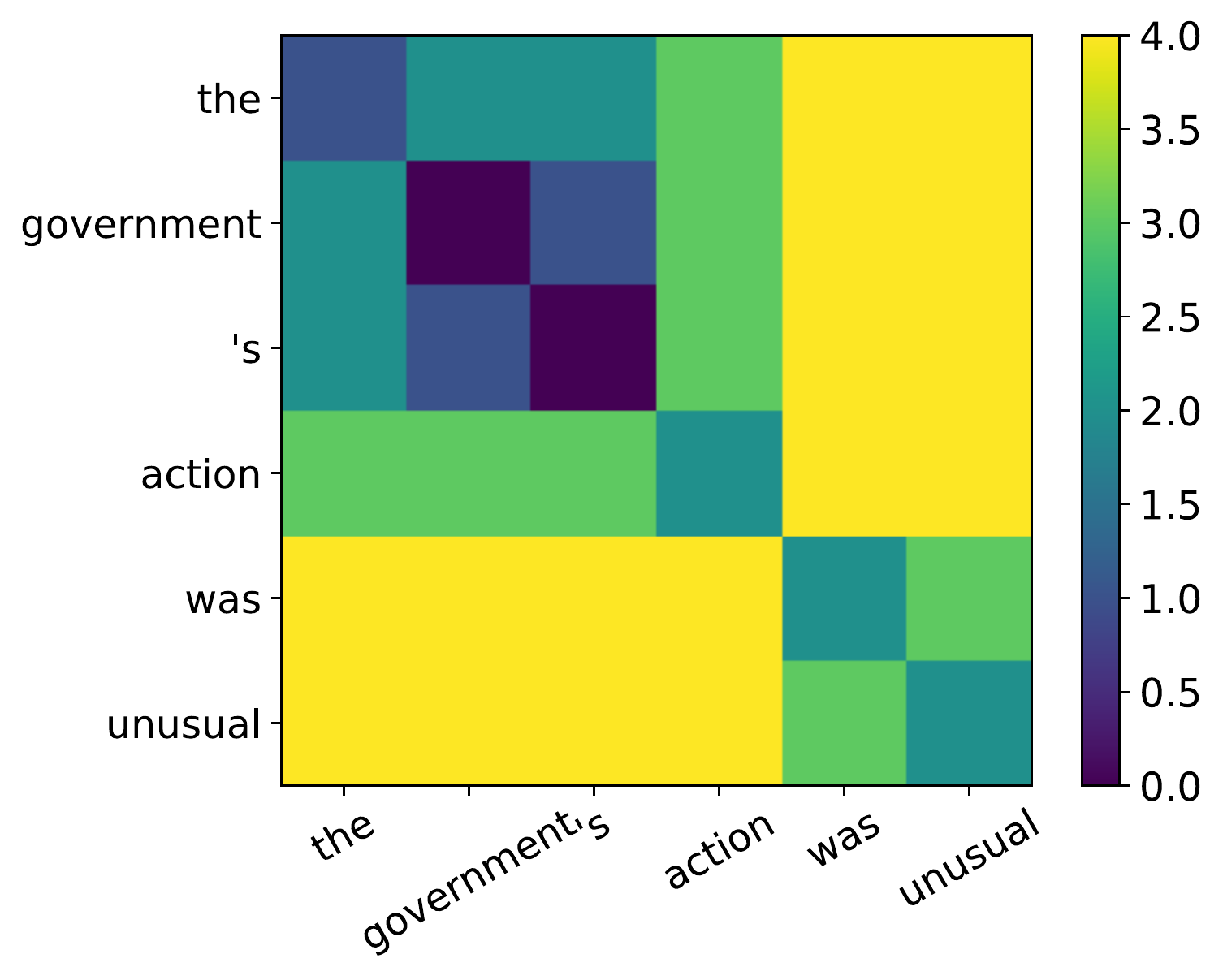} 
	\caption{}
    \label{fig:tree-matrix-b}
	\end{subfigure}
\label{fig:tree-matrix}
\caption{The constituent tree (a) and the distance matrix corresponding to it (b) of an example sentence ``The government 's action was unusual''}
\end{figure}
In this section, we first introduce the syntactic distance presented by \citet{shen2018neural} and propose a new approach to calculating syntactic distance for unsupervised parsing from a distance matrix. 

According to \citet{shen2018neural}, the syntactic distance of two leaves is defined as the height of the lowest common ancestors of these two leaves. For the unlabeled constituent tree in Figure \ref{fig:tree-matrix-a}, the syntactic distance of leaf ``The'' and leaf ``government'' is 2, while the syntactic distance of leaf ``government'' and leaf ``s'' is 1. \citet{shen2018straight} propose a top-down parsing method that splits a span according to the syntactic distance between two adjacent leaves. A span is split between two adjacent words with the largest syntactic distance. The large syntactic distance indicates the two leaves share few common ancestors. For the tree in Figure \ref{fig:tree-matrix-a}, the leaf ``action'' has the largest syntactic distance (4) with its adjacent leaf ``was'' and the smallest syntactic distance (2) with itself. 

The syntactic distance defined in the aforementioned way shares some common properties with the dot product of two vectors. Firstly, syntactic distance measures the dissimilarity between two words according to the number of their common ancestors in the tree, while the dot product can be used to measure the similarity between two words with their vector representations. Secondly, a leaf node has the smallest syntactic distance with itself, while a vector has a high dot product with itself. Therefore we use the dot product of vectors to approximate the negative syntactic distance\footnote{The negative syntactic distance is not a negative number, it is the number of common ancestors of two leaves. The smaller the syntactic distance is, the more ancestors the two leaves share.} of two words. The self-attention matrix of every head of a pretrained language model (particularly BERT used in this paper) contains normalized dot products of different word representations. Therefore every attention matrix can be taken as an approximation to the negative syntactic distance matrix.

If two spans have no intersection, the syntactic distance from any word in one span to any word in the other span is the same. For example in Figure \ref{fig:tree-matrix-b}, the syntactic distances from any word in the span ``The government ’s action'' to any word in the span ``was unusual'' are all 4. Although we can take the negative attention weight of two adjacent words as the syntactic distance and apply the top-down algorithm proposed by \citet{shen2018straight} for unsupervised parsing, it is better to exploit the attention weights between two adjacent spans, instead of adjacent words, to estimate the syntactic distance, which can reduce the estimation bias of of syntactic distance. Given two adjacent spans: $\text{span}(x,y)$, $\text{span}(y+1,z)$, we average the negative attention weights from words in one span to words in the other span as the syntactic distance between these two spans:
\begin{equation}
\label{eq:outside-split-score}
    d(\text{span}_{(x,y)},\text{span}_{(y+1,z)})=-\frac{\sum\limits_{i=x}^{y} \sum\limits_{j=y+1}^{z} a_{ij}
+{\sum\limits_{i=y+1}^{z}\sum\limits_{j=x}^{y}} a_{ij}}
{2(y-x+1)(z-y)}
\end{equation}
where $a_{ij}$ is the attention weight that word $i$ attends to word $j$. We can take the syntactic distance between $\text{span}(x,y)$ and $\text{span}(y+1,z)$ as the split score of split point $y$ in $\text{span}(x,z)$, and parse a constituent tree from a self-attention weight matrix with the greedy algorithm or the chart based algorithm introduced in Section \ref{section:Parsing with Split Score}. 

Although our methods estimating split scores is inspired by the syntactic distance proposed by \citet{shen2018ordered}, trees parsed by our method do not have the right-branching bias, since we estimate the syntactic distance for every two adjacent spans, rather than every individual word, as suggested by \citet{rb_bias}. \citet{Kim2020Are} exploit the hidden states of pretrained language model to estimate the syntactic distance. The key difference between our method and theirs is that our syntactic distance is estimated with the attention weights between two spans rather than two adjacent words.

\subsection{Estimating Split Scores based on the Strength of Inside and Outside Association}
In a constituent tree, intuitively the strength of the association among words inside a constituent (\textbf{inside association}) is stronger than that of association to words outside the constituent (\textbf{outside association}). The syntactic distance can be considered to be related with the outside association, which measures the distance between two adjacent spans. However, it does not consider the inside association, which measures the distance between words inside the span.

\begin{figure}[t]
\setlength{\belowcaptionskip}{-0.4cm}
    \centering
    \includegraphics[width=0.3\textwidth]{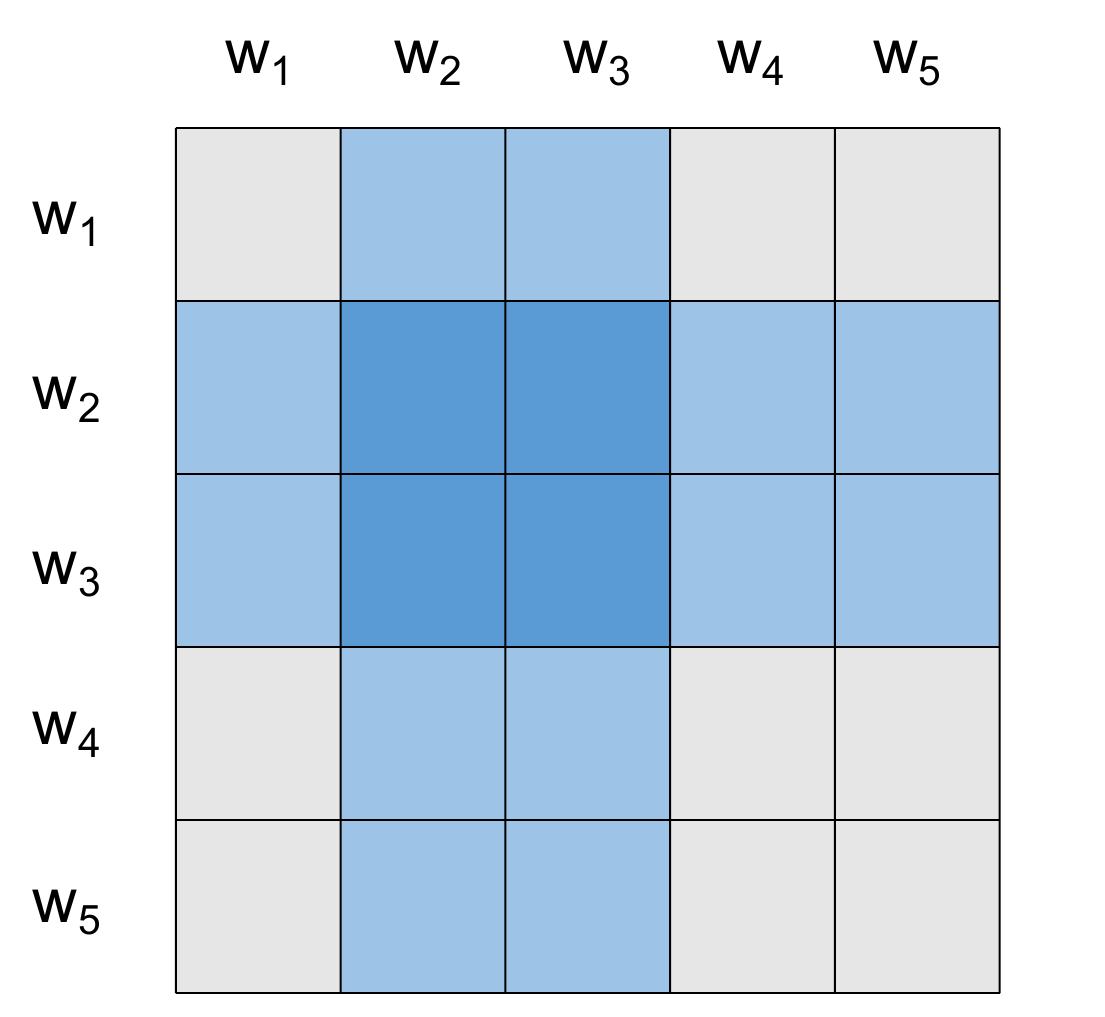}
    \caption{The self-attention weight matrix of an example ``$\text{w}_1$ $\text{w}_2$ $\text{w}_3$ $\text{w}_4$ $\text{w}_5$''. The dark-blue region indicates the attention of span ($\text{w}_2,\text{w}_3$) to itself, while the light-blue region indicates the attention of span ($\text{w}_2,\text{w}_3$) to outside context and the attention of outside context to span ($\text{w}_2,\text{w}_3$).}
    \label{fig:attention}
\end{figure}

Based on the aforementioned intuition, we compute the split score according to the inside association together with outside association. We first define the span score which measures how likely a span can be a constituent. For this, we again resort to the self-attention weights of pretrained language models, as they can be regarded as a relatedness metric between words. Given a span $(x,y)$ from position $x$ to $y$, the stronger the inside association (shown in dark blue in Figure \ref{fig:attention}) and the weaker the outside association (shown in light blue in Figure \ref{fig:attention}), the more possible that span $(x,y)$ is functioning as a constituent. Therefore we estimate two scores for $(x,y)$: $s_{\text{in}}(x,y)$ measuring the strength of the inside association and $s_{\text{out}}(x,y)$ for the outside association. We define $s_{\text{in}}(x,y)$ as the average attention weights $a_{ij}$ inside the span which can be formulated as follows:
\begin{equation}
s_{\text{in}}(x,y)=\frac{\sum_{i=x}^{y} \sum_{j=x}^{y} a_{ij}}{(y-x+1)^2}
\label{inside-rel-score}
\end{equation}
Generally, the span score of a span is not only related to its adjacent span, but all other words outside the span. Therefore we formulate the outside association as the average attention weights between the words inside the span and all other words outside the span:
\begin{equation}
s_{\text{out}}(x,y)=
\frac{{\sum\limits_{i=x}^{y} \sum\limits_{j=0,j\notin[x,y]}^{n-1}} a_{ij}
+{\sum\limits_{i=0,i\notin[x,y]}^{n-1}\sum\limits_{j=x}^{y}} a_{ij}}
{2(y-x+1)n-2(y-x+1)^2}
\label{inside-outside-rel-score}
\end{equation}
where $n$ is the length of the sentence that consumes the span $(x,y)$. The numerator of the above equation is the sum of all attention weights $a_{ij}$ inside the four sub-matrices corresponding to the outside association, while the denominator is the sum of the area of the four sub-matrices. The first component in the numerator estimates the attention association from words inside the span to words outside the span while the second indicates the attention association from words outside the span to words inside the span. Please notice that these two components are not equal to each other as the self-attention weight matrix is not symmetric. 

Given the inside association score $s_\text{in}(x,y)$ and the outside association score $s_\text{out}(x,y)$, the score of span $(x,y)$ to be a constituent can be estimated as follows:
\begin{equation}
s_{\text{span}}(x,y)=s_{\text{in}}(x,y)-s_{\text{out}}(x,y)
\label{span-score}
\end{equation}
Given the span score of two adjacent spans: $s_{\text{span}}(x,z-1)$ and $s_{\text{span}}(z,y)$, we can estimate the split score as the sum of two span scores:
\begin{equation}
\label{eq:in-out-split-score}
s_{\text{split}}(x,y,z)=s_{\text{span}}(x,z-1)+s_{\text{span}}(z,y)
\end{equation}

The differences between the split score based on the syntactic distance estimated in Eq. (\ref{eq:outside-split-score}) and that according to the strength of the inside and outside association in Eq. (\ref{eq:in-out-split-score}) are twofold. First, the syntactic distance is just corresponding to the strength of the outside association between two adjacent spans. Second, the syntactic distance computed in Eq. (\ref{eq:outside-split-score}) is the negative average attention weights between words in two adjacent spans, while the outside association in this section is the average attention weights between words inside a span and all other words outside the span. 

\subsection{Parsing with Split Scores } \label{section:Parsing with Split Score}
\paragraph{Greedy parsing.} The inference of a tree from a sentence can be viewed as a process of splitting the sentence recursively until the spans can not be split any more (i.e., length-one spans). For a span with length $m$, there are $m-1$ options to split this span into two sub-spans. 

Given a span $(i,j)$, we split the span at the split point with the highest split score:
\begin{equation}
k^*=\argmax_{k=i+1,...,j}(s_{\text{split}}(i,j,k))\label{eq:greedy-split-point}
\end{equation}
where $s_{\text{split}}(i,j,k)$ is the split score of span (i,j) at split point k. When splitting a span into two sub-spans, we only consider the largest split score of the span, ignoring the split score of the sub-spans. Therefore this parsing algorithm is greedy, the time complexity of which is $O(n^2)$

\paragraph{Chart parsing.} To incorporate split scores of sub-spans into the splitting decision of a current span, we define an enhanced split score inspired by \citet{stern-etal-2017-minimal}:
\begin{equation}
\tilde{s}_{\text{split}}(i, j, k)=s_{\text{split}}(i, j, k) + s_{\text{best}}(i, k-1)+s_{\text{best}}(k, j)
\end{equation}
where $s_{\text{best}}(i,k-1)$ and $s_{\text{best}}(k,j)$ are the best span score of span $(i,k-1)$ and span $(k,j)$. Given a span $(i,j)$, we can compute the best span score and the best split point of it with the augmented split score as follows: 
\begin{equation}
\begin{split}
s_{\text{best}}(i,j)=\max_{k=i+1,...,j}(\tilde{s}_{\text{split}}(i,j,k)) \\
k^*=\argmax_{k=i+1,...,j}(\tilde{s}_{\text{split}}(i,j,k))
\label{best-split-score}
\end{split}
\end{equation}
We compute the best span score and the best split point for all spans of a given sentence bottom up according to Eq. (\ref{best-split-score}). During this process, we store the best split points in a chart. After that, we call a top-down parsing procedure with the best split points kept in the chart in a recursive manner to infer a binary tree. Our chart parsing is actually a CKY-style parsing\footnote{\url{https://en.wikipedia.org/wiki/CYK_algorithm.}}, which has the $O(n^3)$ time complexity.

\section{Few-Shot Parsing}\label{section:Few-Shot Parsing}
\begin{figure}[t]
\centering
   \begin{subfigure}[b]{0.3\linewidth}	
   \centering
	\includegraphics[width=\linewidth]{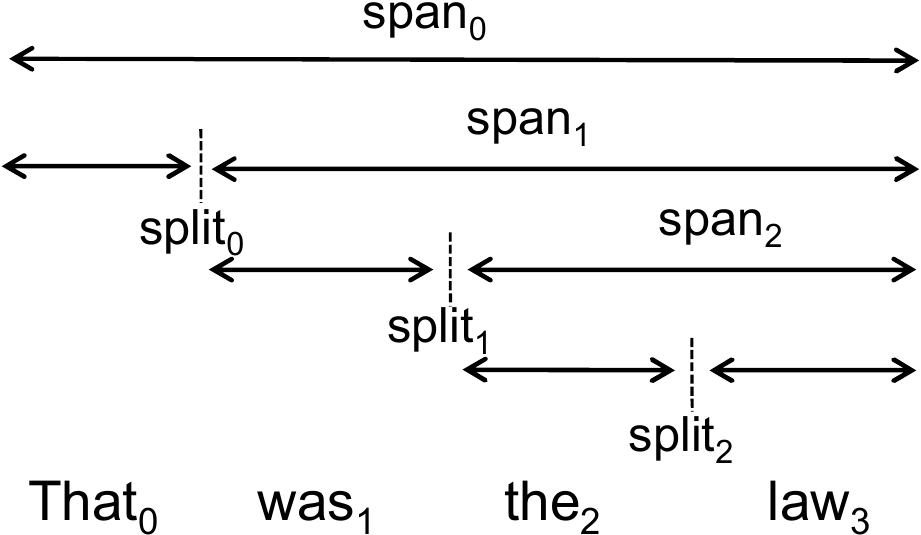} 
	\caption{}
\label{fig:tree2split-a}
	\end{subfigure}
    \quad\quad\quad\quad
   \begin{subfigure}[b]{0.3\linewidth}	
     \centering
	\includegraphics[width=\linewidth]{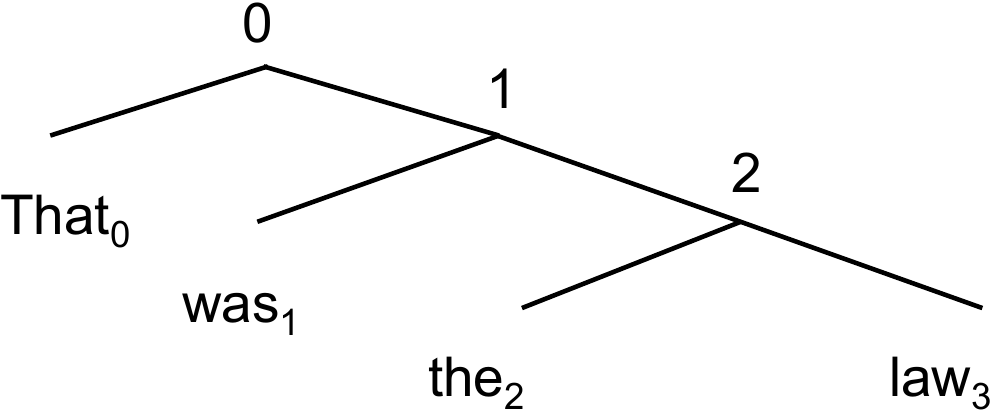}
	\setlength{\abovecaptionskip}{13pt plus 3pt minus 2pt}
	\caption{}
\label{fig:tree2split-b}
	\end{subfigure}
    \caption{An example sentence ``That was the law'' (a) and its corresponding tree (b).}
    \label{fig:tree2split}
\end{figure}
The attention weight matrix of BERT is the normalized dot product of a query $\mathbf{Q}$ and a key $\mathbf{K}$, as shown in Eq. (\ref{attention}). The essential information of the query and key is from the hidden representation $\mathbf{H}_{L-1}$ of the previous layer. They are computed as the product of $\mathbf{H}_{L-1}$ with $\mathbf{W}_Q $ and $\mathbf{W}_K$, as shown in Eq. (\ref{projection}). We denote the dimension of $\mathbf{H}_{L-1}$ as $d_{\text{model}}$. $\mathbf{W}_{Q},\mathbf{W}_{K} \in R^{d_{\text{model}} \times d_{\text{model}}/h}$ , $h$ is the number of attention heads. BERT uses two matrices ($\mathbf{W}_Q, \mathbf{W}_K$) to linearly project $\mathbf{H}_{L-1}$ onto different sub-spaces and then computes the attention matrix as follows: 
\begin{equation}
Attention=\softmax(\frac{\mathbf{Q}\mathbf{K}^T}{\sqrt{d_{\text{model}}}})
\label{attention}
\end{equation}

\begin{equation}
\begin{split}
\mathbf{Q}=\mathbf{H}_{L-1}\mathbf{W}_Q\\
\mathbf{K}=\mathbf{H}_{L-1}\mathbf{W}_K
\end{split}
\label{projection}
\end{equation}

We have explored all attention weight matrices of BERT to unsupervisedly parse sentences as described in Section \ref{section:Unsupervised parsing with different heads}, and have found that the parsing results of different heads vary greatly even in the same layer. The differences among different heads are essentially from the different two linear projections: $\mathbf{W}_Q$ and $\mathbf{W}_K$. Therefore, we conjecture that they have a great impact on parsing and that we may infer better parse trees with better projections. 

To verify this, we freeze other parameters of BERT, and retrain the two linear projections ($\mathbf{W}_Q,\mathbf{W}_K$) with annotated trees. Given the binarized tree of a sentence of length $n$, there are $n-1$ split points in the tree, and each split point is corresponding to a span of the sentence and an internal node of the tree as shown in Figure \ref{fig:tree2split}. We denote the $j$th split point as $\text{split}_j$ and the span split by $\text{split}_j$ as $\text{span}_j$.  For example, $\text{split}_1$ in Figure \ref{fig:tree2split-a} is corresponding to the $\text{span}_1$ (i.e., (1,3)) and the internal node 1 in Figure \ref{fig:tree2split-b}. We can compute the split scores of these split points with Eq. (\ref{eq:outside-split-score}) or Eq. (\ref{eq:in-out-split-score}). To make the computation suitable to our training, we normalize the split scores of each span using the softmax function:
\begin{equation}
p(\text{split}_j\mid \text{span}_j;\theta)=\frac{e^{s_\text{split}(\text{span}_j, \text{split}_j)}}{\sum_{k=1}^{n-1}e^{s_\text{split}(\text{span}_j,k)}}
\label{probability-of-split}
\end{equation}
where $\theta$ is the two matrices $\mathbf{W}_Q$, $\mathbf{W}_K$, $s_\text{split}(\text{span}_j, k)$ is defined in Eq. (\ref{eq:outside-split-score}) or Eq. (\ref{eq:in-out-split-score}). The normalized scores can be seen as the probability of choosing a split point from the corresponding span. Multiplying the probabilities of all split points in a tree, we can get the probability of this tree. 
Given a binarized ground-truth tree, which contains $n-1$ split points, we can compute the probability of the tree as follows: 
\begin{equation}
p(\text{tree}\mid S;\theta)=\prod_{\text{split}_j\in \text{tree}}{p(\text{split}_j\mid \text{span}_j;\theta)}
\label{probability-of-tree}
\end{equation}
where $\text{split}_j \in \text{tree}$ denotes that $\text{split}_j$ is corresponding to a node in the tree and $S$ represents the sentence. We maximize the probability of ground-truth trees to optimize $\theta$, which is equivalent to minimizing the negative log likelihood:
\begin{equation}
\label{eq:mle-loss}
L_{\theta}^{\text{MLE}}=-\sum_i^N log(p(\text{tree}_i\mid S_i; \theta))
\end{equation}
where $N$ is the number of annotated trees used for training $\theta$, $\text{tree}_i$ is the binarized annotated tree of sentence $S_i$. Minimizing the negative log likelihood forces the model to maximize the probability of the ground-truth trees, which may cause over-fitting on the small training set. Actually, as long as the scores of the ground-truth trees are higher than other trees, we can derive the ground truth trees. Therefore we define a margin loss as the new training loss:
\begin{equation}
\label{eq:margin-loss}
L_{\theta}^{\text{Margin}}=\sum_i^N \sum_{\text{span}_j\in \text{tree}_i} \sum_{\text{split}_k\in \text{span}_j} max(0,\text{margin}+s_\text{{split}}(\text{span}_j,\text{split}_k)-s_\text{{split}}(\text{span}_j,\text{split}_{k^*}))
\end{equation}
Where the $\text{span}_j$ is a span in the $\text{tree}_i$, and $\text{split}_k$ is a split point in the $\text{span}_j$. $\text{split}_{k^*}$ is a gold split point that appears in the ground truth tree. The computation of both the negative likelihood and the margin loss are differentiable, as equations associated with them (Eq. (\ref{eq:outside-split-score}), (\ref{inside-rel-score}), (\ref{inside-outside-rel-score}), (\ref{span-score}), (\ref{eq:in-out-split-score}), (\ref{attention}), (\ref{projection}), (\ref{probability-of-split}), (\ref{probability-of-tree})) are all differentiable. Therefore the model can be optimized by gradient descent. Since  the parameters to be tuned are just $\mathbf{W}_Q$ and $\mathbf{W}_K$, the two linear projection matrices, we can use a few tree samples to well train them. 

 For inference, we use the trained projection matrices to replace the original ones in BERT and yield parse trees via the parsing algorithm in Section \ref{section:Parsing with Split Score}. 
\section{Experiments}
We carried out two groups of experiments to evaluate the proposed unsupervised and few-shot parsing models. One is a series of unsupervised and few-shot parsing experiments on the English treebank PTB \citep{ptb}. The other is a group of cross-lingual parsing experiments, where the unsupervised model and few-shot model tuned/trained on PTB were evaluated on 8 treebanks of different languages.
\begin{table}
    \caption{\label{table:dataset size} The dataset size of the treebank of 9 languages. }
    \centering
    \begin{tabular}{c l l l l l l l l l}
    \toprule 
    \textbf{dataset} & \textbf{English} & \textbf{Korean} & \textbf{German} & \textbf{Polish} & \textbf{Hungarian} & \textbf{Basque} & \textbf{French} & \textbf{Hebrew} & \textbf{Swedish} \\
    \midrule
        training & 45,089 & 23,010 & 40,472 & 6,578 &  8,146 &  7,577 & 14,759 & 5,000 &  5,000  \\
        validation & 1,700  & 5,000 & 821 & 3,059 & 1,051 & 948 &  1,235 & 500 & 494  \\
        test & 2,416 & 2,287 & 5,000 & 822 & 1,009 & 946 & 2,541 & 716 & 666  \\
    \bottomrule
    \end{tabular}
    \label{tab:data-szie}
\end{table}
\subsection{Experiment Setting}
We evaluated our unsupervised and few-shot parser on PTB \citep{ptb} for English and SPMRL \citep{spmrl} for eight languages. We used the standard split of PTB and SPMRL. The sizes of train/dev/test data of all languages are shown in Table \ref{tab:data-szie}. Following \citet{shen2018ordered}, we removed the punctuation in PTB and used evalb\footnote{Available at https://nlp.cs.nyu.edu/evalb/.} to evaluate the performance of parsing. As the F1 score from evalb is computed at the corpus level, we also report the sentence-level F1 score for the comparison with \citet{Kim2020Are}. 

We adopted the word piece method proposed by \citet{devlin-etal-2019-bert} for tokenization, where words were tokenized into multiple sub-words. To match our parsed trees with the gold-standard trees, we need to merge the attention weights of sub-words in the attention weight matrix. For attention to a split-up word, we summed up the attention weights to all word pieces of the word as the attention weights to the word. For attention from a split-up word, we take the mean of the attention weights from word pieces of the word as the attention weights from the word. 

For experiments on PTB, our models were built based on the pretrained base-uncased BERT released by \citet{devlin-etal-2019-bert}. In unsupervised experiments on PTB, the raw texts of the training set were used for training, while the annotated trees of the validation set were used for tuning hyper-parameters. We compared our unsupervised parsing methods against previous state-of-the-art unsupervised models\footnote{We did not compare our method with DIORA \citep{drozdov-etal-2019-unsupervised} and URNNG \citep{kim-etal-2019-unsupervised}, because these two methods are evaluated on sentences without removing punctuation.} including PRPN \citep{shen2018neural}, Ordered neurons \citep{shen2018ordered}, Tree-Transformer \citep{Chen-2019-tree}, C-PCFG \citep{kim-etal-2019-compound}; \citet{Kim2020Are}, which is also built on the attention matrix of BERT. For few-shot experiments on PTB, only a few randomly chosen trees from the training set were used for few-shot training. We compared our few-shot parsing methods with a few-shot baseline \citep{shi2020role} and Berkeley parser \citep{kitaev-etal-2019-multilingual}, which is a state-of-the-art supervised parser.

For experiments on SPMRL, our models were built based on the pretrained multilingual-base-uncased BERT released by \citet{devlin-etal-2019-bert}.  The experiments on SPMRL were cross-lingual, where the unsupervised and few-shot parser tuned/trained on the PTB were evaluated on the test data in SPMRL. A multi-lingual unsupervised parser \citep{kimchart} and the multi-lingual Berkeley parser were trained and evaluated in the same setting as our methods for comparison.

We denote our unsupervised parser based on the strength of both the inside and outside association as \textbf{UPIO}, the unsupervised parser based on the syntactic distance (outside association) as \textbf{UPOA}, the few-shot parser based on inside and outside association as the \textbf{FPIO}, and the few-shot parser based on the syntactic distance (outside association) as the \textbf{FPOA}.

\subsection{Unsupervised Parsing with Different Heads on PTB}\label{section:Unsupervised parsing with different heads}
\begin{figure}[t] 
    \centering
    \begin{subfigure}[b]{0.4\linewidth}	
        \centering
        \includegraphics[width=1.0\linewidth]{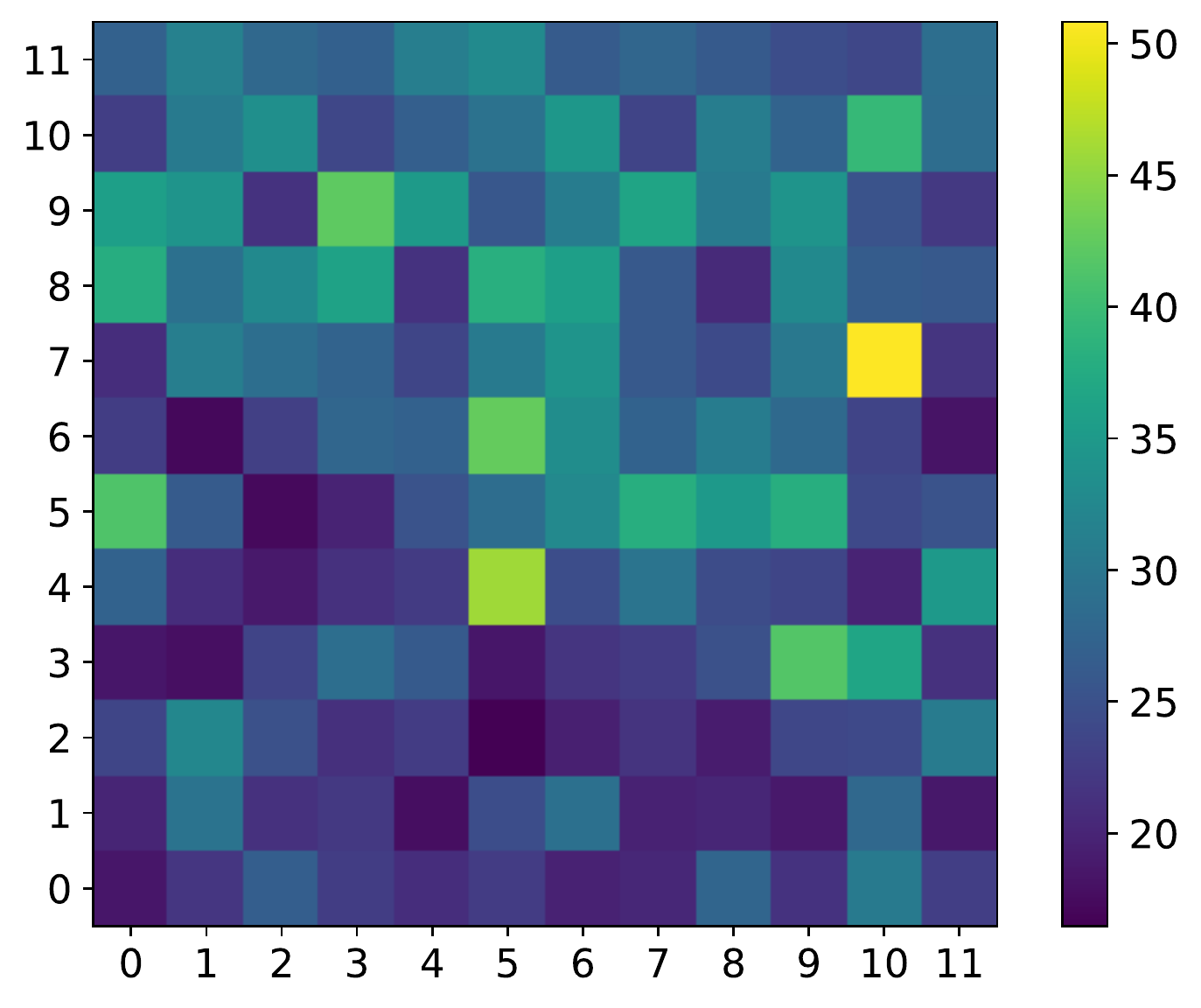} 
        \caption{}
        \label{fig:parse-on-diff-heads-b}
    \end{subfigure}
    \begin{subfigure}[b]{0.4\linewidth}	
        \centering
        \includegraphics[width=1.0\linewidth]{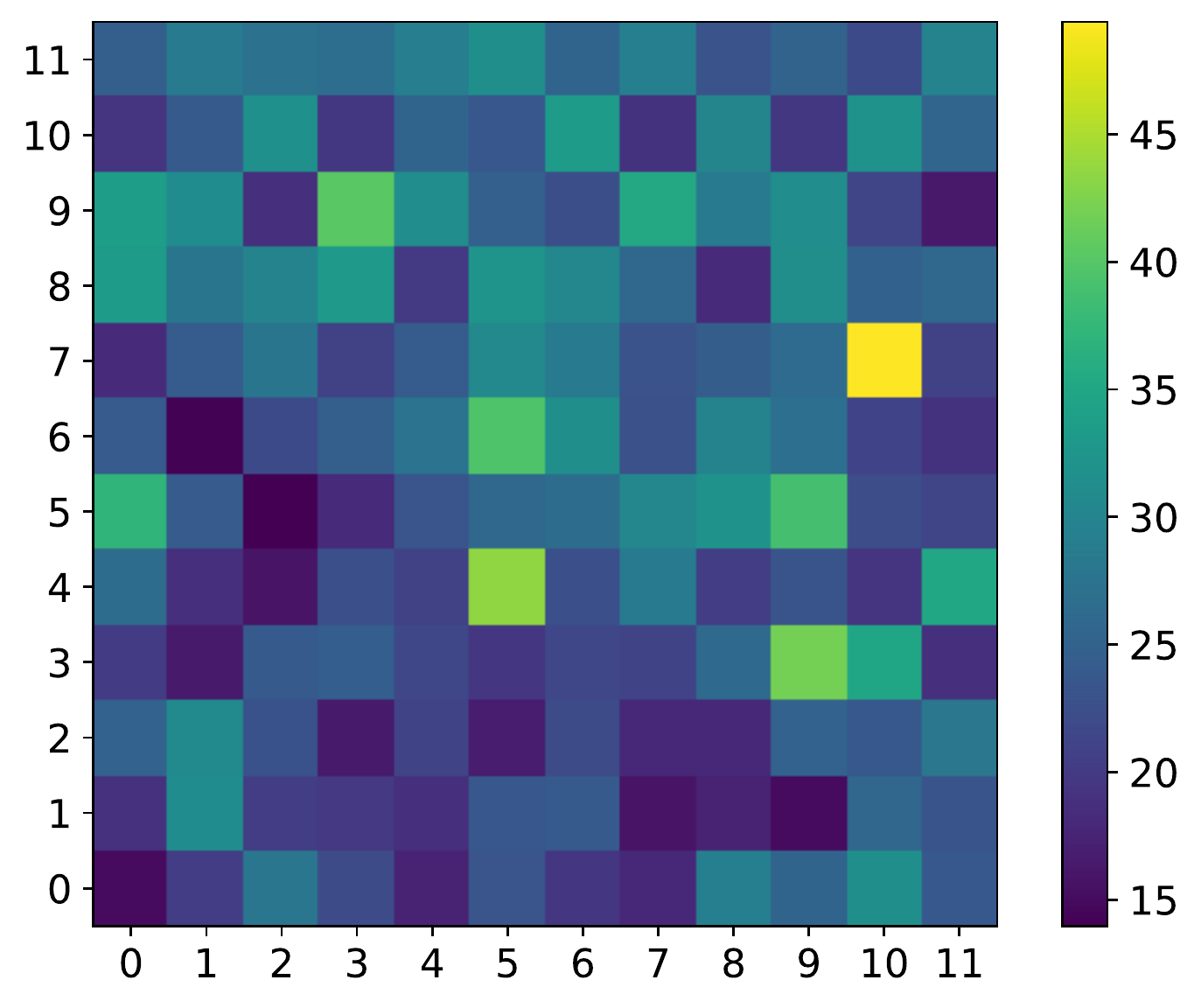} 
        \caption{}
        \label{fig:parse-on-diff-heads-a}
    \end{subfigure}
    \caption{Unsupervised parsing performance of UPOA (a) and UPIO (b) with different attention heads on validation set. The pixel in the $i$th row, and $j$th column indicates that the F1-score of UPOA and UPIO with the $j$th head of $i$th layer. The lighter the color, the higher the F1 score. }
    \label{fig:parse-on-diff-heads}
\end{figure}
There are 12 layers in BERT base model, and each layer contains 12 attention heads. We tried all 12$\times$12 heads of BERT without fine-tuning BERT on PTB raw texts to produce syntactic trees with the proposed UPOA and UPIO. In Section \ref{section:Parsing with Split Score}, we introduce two parsing algorithms, the chart-based algorithm considers the split scores of sub-spans, which is helpful for the UPIO but not for UPOA. Therefore, all experiments results from UPIO were obtained with the chart parsing algorithm, while all experiments results from UPOA were obtained with the greedy parsing algorithm. The relationship between the parsing algorithm and the parsing performance of UPOA and UPIO will be analyzed in Section \ref{section:Ablation study}.

Results are shown in Figure \ref{fig:parse-on-diff-heads}. The color of Figure \ref{fig:parse-on-diff-heads-a} is very similar to that of Figure \ref{fig:parse-on-diff-heads-b} at any positions of the matrix, indicating that each head for UPIO has the similar performance to that of the same head for UPOA. However, the color map in Figure \ref{fig:parse-on-diff-heads-a} is different from that in Figure \ref{fig:parse-on-diff-heads-b}. The performances for most heads on UPOA are better than the same heads for UPIO, suggesting that UPIO always get similar or worse results compared with UPOA, although the split score in UPIO exploit more information (inside association) than UPOA. We will show that UPIO is superior to UPOA on short sentences in Section \ref{section:Unsupervised parsing results on PTB} and that the few-shot version of UPIO (i.e., FPIO) trained with a few annotated trees performs better than that of UPOA (FPOA) in Section \ref{section:Few-shot parsing results on PTB}. Therefore, the inside association is helpful for parsing, but it can not be estimated exactly with the original attention weights on long sentences.

From either Figure \ref{fig:parse-on-diff-heads-b} or Figure \ref{fig:parse-on-diff-heads-a}, we can observe that attention heads in the higher layers perform significantly better than those in lower layers. The F1 score of different attention heads varies greatly even in the same layer. \citet{clark-etal-2019-bert} also draw a similar conclusion that only a few heads obtain good results on predicting syntactic relations. The only difference among attention heads in the same layer is from the corresponding linear projections, which trigger the varying parsing results with different heads in the same layer. Such finding has directly inspired us to train better linear projections in the few-shot parsing framework. 
 
\subsection{Unsupervised Parsing Results on PTB}\label{section:Unsupervised parsing results on PTB}
\begin{table}[t]

\centering
\caption{\label{parsing-on-wsj}F1 scores of unsupervised parsing on PTB. WSJ-test denotes the test set of PTB, while WSJ10 is a subset of the whole PTB where sentence length is <= 10. Tag accuracy is the phrase detection accuracy on different syntactic categories. The results shown in the top half of the table are evaluated with corpus-level F1 score, while those of the bottom half of the table are evaluated with sentence-level F1 score. The results of PRPN and ON-LSTM are from \citet{shen2018ordered}. Tree-T: Tree-Transformer\citep{Chen-2019-tree}. \dag: without fine-tuning on annotated trees from the PTB validation set. \ddag: without fine-tuning on the raw texts of the PTB training set. LB: left branching. RB: right branching.}
\begin{tabular}{c c c c c c c c c c c c}
\toprule 
\multirow{3}{*}{\textbf{Model}} & \multicolumn{6}{c}{\textbf{Corpus level F1 score}} & \multicolumn{5}{c}{} \\
\cmidrule(lr){2-7}
& \multicolumn{3}{c}{\textbf{WSJ-test}} & \multicolumn{3}{c}{\textbf{WSJ10}} & \multicolumn{5}{c}{\textbf{Tag Accuracy}} \\
\cmidrule(lr){2-4} \cmidrule(lr){5-7} \cmidrule(lr){8-12}
& \textbf{$\mu$} &\textbf{$\sigma$} & \textbf{max} & \textbf{$\mu$} & \textbf{$\sigma$} & \textbf{max} & \textbf{NP} & \textbf{VP} & \textbf{PP} & \textbf{ADJP} & \textbf{SBAR} \\ \midrule
ON-LSTM & 47.7 & 1.6 &  49.4 &  65.1 & 1.7 & 66.8 & 71.2 & 33.8 & 58.8 & 32.5 & 52.5 \\
Tree-T & 49.5 & -  & 51.1 & 66.2 & - & 68.0 & 67.6 & 38.5 & 52.3 & 24.7 & 36.4 \\
C-PCFG & \textbf{52.4} & - & - &  - & - & - & \textbf{74.7} & \textbf{41.7} &  68.8 & 40.4 & \textbf{56.1}\\
UPOA & 50.9 & 0.2 & \textbf{51.4} & 72.6 & 0.6 & 73.8 & 65.5 & 34.2 & \textbf{79.7} & 45.6 & 55.4 \\
UPIO & 50.3 & 0.4 & 51.0 & \textbf{73.8} & 0.4 & 74.5 & 64.1 & 37.7 & 76.4 & \textbf{50.2} & 46.5\\\midrule
PRPN\textsuperscript{\dag} & 37.4 & 0.3 & 38.1 & 70.5 & 0.4 & \textbf{71.3} & \textbf{63.9} & - & 24.4 & 26.2 & - \\
UPOA\textsuperscript{\dag} & 40.1 & 0.5 & 40.9 & - & - & 62.5 & 57.6 & 24.5 & 39.7 & 50.5 & \textbf{38.9} \\
UPIO\textsuperscript{\dag} & \textbf{42.2} & 0.2 & \textbf{42.4} & - & - & 64.5 & 57.4 & 26.9 & \textbf{47.1} & \textbf{53.4} & 36.3 \\
Random & - & 21.6 & - &  21.8 & 31.9 & 32.6 & - & - & - & - & -  \\
LB & - & 9.0 & - &  9.0 & 19.6 & 19.6 & - & - & - & - & - \\
RB & - & 39.8 & - &  39.8 & 56.6 & 56.6 & - & - & - & - & -  \\

& \multicolumn{6}{c}{\textbf{Sentence-level F1 score}} & \multicolumn{5}{c}{\textbf{Tag Accuracy}} \\
\cmidrule(lr){2-7} \cmidrule(lr){8-12}
\citet{Kim2020Are}\textsuperscript{\ddag} & 42.3 & 0 & 42.3 & - & - & - & 46.0 & \textbf{49.0} & 43.0 & 41.0 & \textbf{65.0} \\
UPOA\textsuperscript{\ddag} & \textbf{50.9} & 0 & \textbf{50.9} & 72.9 & 0 & 72.9 & \textbf{65.6} & 33.2 & \textbf{78.5} & \textbf{43.8} & 59.6 \\
UPIO\textsuperscript{\ddag} & 50.4 & 0 & 50.4 & \textbf{74.5} & 0 & \textbf{74.5} & 63.6 & 36.4 & 75.8 & \textbf{43.8} & 48.5 \\\midrule
\citet{heads_up}\textsuperscript{\dag\ddag} & \textbf{42.7} & 0 & \textbf{42.7}  & - & - & - & 49.0 & \textbf{30.0} & 42.0 & 40.0 & \textbf{69.0} \\
UPOA\textsuperscript{\dag\ddag} & 39.6 & 0 & 39.6 & 56.1 & 0 & 56.1 & \textbf{55.1} & 25.1 & 39.0 & 46.7 & 36.1 \\
UPIO\textsuperscript{\dag\ddag} & 39.2 & 0 & 40.5 & \textbf{56.9} & 0 & \textbf{56.9} & 54.0 & 25.9 & \textbf{46.2} & \textbf{48.5} & 34.2 \\
\bottomrule
\end{tabular}
\end{table}

\begin{table}[t]

    \centering
    \caption{\label{parsing-on-subsets-of-wsj}F1 scores and parsing speed of unsupervised parsing on 5 subsets of WSJ-test. ``10'',``20'',``30'',``40'' denote the subset of WSJ-test where sentence length is $<=$ 10,20,30,40 respectively. ``all'' denotes all sentences in WSJ-test. The parsing speed is measured by the number of processed sentences per second.}
    \begin{tabular}{ccccccccccc}
    \toprule
        \multirow{2}{*}{Model} & \multicolumn{2}{c}{10} & \multicolumn{2}{c}{20} & \multicolumn{2}{c}{30} & \multicolumn{2}{c}{40} & \multicolumn{2}{c}{all} \\
        \cmidrule(lr){2-3} \cmidrule(lr){4-5} \cmidrule(lr){6-7} \cmidrule(lr){8-9} \cmidrule(lr){10-11}
        & F1 & Speed & F1 & Speed & F1 & Speed & F1 & Speed & F1 & Speed \\
        \midrule
        UPIO & 69.4 & 23.4 & 57.6 & 3.6 & 52.7 & 1.5 & 50.9 & 0.9 & 50.3 & 0.8 \\
        UPOA & 67.7 & 66.3 & 57.2 & 16.3 & 53.1 & 6.8 & 51.4 & 4.8 & 50.9 & 3.5 \\
        C-PCFG& 70.6 & 14.0 & 62.0 & 3.9 & 57.1 & 1.73 & 54.6 & 1.2 & 52.4 & 0.8 \\
    \bottomrule
    \end{tabular}
    \label{tab:my_label}
\end{table}

In this section, we compared our proposed UPOA and UPIO against other unsupervised methods on PTB. The results are shown in Table \ref{parsing-on-wsj}. The baselines in the top half of Table \ref{parsing-on-wsj} were evaluated with the corpus-level F1 score\footnote{Corpus-level F1 score measures the precision and recall at the corpus level}, while those in the bottom half of Table \ref{parsing-on-wsj} were evaluated with sentence-level F1 score.\footnote{Sentence-level F1 score is the average F1 score over all sentences.} PRPN \citep{shen2018neural} and \citet{heads_up} are fully unsupervised parsers \footnote{ON-LSTM (Ordered Neurons) needs annotated trees to select the best layer for parsing, hence we do not take it as a fully unsupervised parser}, while other baselines tune their hyper-parameters on annotated trees of the validation set. \citet{Kim2020Are} and \citet{heads_up} are built on the original BERT, while other baselines are trained on the raw texts of the training set. For a fair comparison with them, we report results of 4 versions of our methods:
\begin{enumerate}
\item tuning hyper-parameters on the validation set, fine-tuning on the raw texts of the training set (UPOA, UPIO)
\item fully unsupervised, fine-tuning on the raw texts of the training set  (UPOA\textsuperscript{\dag}, UPIO\textsuperscript{\dag})
\item tuning hyper-parameters on the validation set, built on original BERT (UPOA\textsuperscript{\ddag}, UPIO\textsuperscript{\ddag})
\item fully unsupervised, built on original BERT (UPOA\textsuperscript{\dag\ddag}, UPIO\textsuperscript{\dag\ddag})
\end{enumerate}

The hyper-parameter tuning of UPOA and UPIO is actually to decide which attention head in BERT is to be used for parsing. In Section \ref{section:Unsupervised parsing with different heads}, We have found 3 heads (the 10th head in the 7th layer, the 3th head in the 9th layer and the 5th head in the 4th layer) which significantly outperform other heads in both UPOA and UPIO. We average the attention weights of these three heads in UPOA\textsuperscript{\dag} and UPIO\textsuperscript{\dag}, which can be considered as tuning hyper-parameters on the validation set. For the fully unsupervised version of UPOA and UPIO, we simply average all attention heads in BERT. The fine-tuning of UPOA and UPIO is just fine-tuning BERT on the raw texts of the training set of PTB with the masked language model objective for 10 epochs, following the setting of \citet{devlin-etal-2019-bert}. Inspired by \citet{shen2018ordered}, we fine-tune BERT with five different seeds and report the average F1 score and maximum F1 score from the 5 runs. The effect of fine-tuning and hyper-parameter tuning will be analyzed in Section \ref{section:Ablation study}.

As shown in Table \ref{parsing-on-wsj}, both UPOA and UPIO outperform baselines on WSJ10, including On-LSTM \citep{shen2018ordered} and Tree-Transformer \citep{Chen-2019-tree}, in terms of corpus-level F1 score. Fully unsupervised UPOA\textsuperscript{\dag} and UPIO\textsuperscript{\dag} are better and more stable than PRPN on WSJ-test. UPOA\textsuperscript{\ddag} and UPOA\textsuperscript{\ddag} built on original BERT, are substantially better than the model proposed by \citet{Kim2020Are}, which is also built on the attention matrix of BERT, in terms of sentence-level F1 score. However, UPOA\textsuperscript{\dag\ddag} and UPIO\textsuperscript{\dag\ddag}, which are fully unsupervised and built on original BERT underperforms the model proposed by \citet{heads_up}. We also observe that hyper-parameter tuning (attention head selecting) has a great impact on the performance of UPOA and UPIO. But do we really need the entire validation set to select heads for UPOA and UPIO? We investigate this question in Section \ref{section:unsupervised parsing with different data size}. According to the analysis in Section \ref{section:unsupervised parsing with different data size}, only 1 tree is enough to tune UPOA and UPIO well. Although the model proposed by  \citet{heads_up} is better than the fully unsupervised version of UPOA and UPIO, our methods are able to achieve substantially better results with only 1 annotated tree.

The state-of-the-art C-PCFG has not provided results on WSJ-10. They report their results on 4 subsets of WSJ-test. Therefore, we compared our UPOA and UPIO with C-PCFG on these 4 subsets in Table \ref{parsing-on-subsets-of-wsj} to investigate the performance of our methods on short sentences. The F1 scores reported in Table 3 are averaged over 5 runs with random seeds. We also display the parsing speed, in terms of the number of processed sentences per second, in Table \ref{parsing-on-subsets-of-wsj}. All models are run on a Tesla P100.Please note that WSJ-10 data in Table \ref{parsing-on-wsj} is a subset of the whole PTB, rather than a subset of WSJ-test. Therefore the F1 scores of UPIO and UPOA on sentences with length $<=$ 10 in Table \ref{parsing-on-wsj} and Table \ref{parsing-on-subsets-of-wsj} are different. 

It can be seen that UPIO and UPOA underperform C-PCFG on all subsets of WSJ-test in Table \ref{parsing-on-subsets-of-wsj}. However, the performance gap between UPIO and C-PCFG is very small on sentences of length $<=$ 10, which indicates that our UPIO is competitive to C-PCFG in short sentence (length $<=$ 10) parsing. It is interesting that UPIO outperforms UPOA on short sentences with length $<=$ 20, but underperforms UPOA on long sentences. Since the main difference between UPOA and UPIO is the exploitation of inside association, the reason why UPOA is better than UPIO on long sentences may that inside association scores estimated from large attention matrices for long sentences are noisy.

Although UPOA is not better than C-CPCFG, in terms of F1 score, it is much faster (4x) than UPIO and C-PCFG in all sentence length settings, due to the top-down parsing algorithm, which has $O(n^2)$ time complexity ($n$ is the sequence length). By contrast, both UPIO and C-PCFG use chart parsing, which has $O(n^3)$ time complexity. We also observe that UPIO is faster than C-PCFG on the sentences of length <= 10, but slower on longer sentences, which may be due to the self-attention mechanism that has $O(n^2)$ time complexity.

Following previous works, we also report the phrase detection performance on different syntactic categories of WSJ-test in Table \ref{parsing-on-wsj}. It is interesting to find that both our UPOA and UPIO always achieves the state of the art on PP and ADJP. We manually studied the attention heat-maps visualized from the attention weight matrices used in UPSA and found that ADJPs are usually short phrases and frequently attend to corresponding nouns which they adjoin to. This may make them easily detectable for UPOA and UPIO.

\subsection{Few-Shot Parsing on PTB}\label{section:Few-shot parsing results on PTB}
In this group of experiments, we frozen other parameters of BERT, and only retrained the two linear projections $\mathbf{W}_Q$ and $\mathbf{W}_K$ for FPOA and FPIO. The multi-head attention mechanism of BERT projects the hidden representations from previous layer to a new subspace. To help our model converge faster, we initialized the parameters of linear projections in FPOA \& FPOA with those of the linear projections of corresponding layer in BERT. However, as the dimension of projected embedding in BERT is $d_{\text{model}}/h$ ($h$ is the number of heads) while the dimension of projected embedding in FPOA \& FPOA is $d_{\text{model}}$, there is a mismatch between the linear projection matrices of our FPOA \& FPOA and BERT ($d_{\text{model}}\times d_{\text{model}} \text{ vs. } d_{\text{model}}\times d_\text{model/h}$). For this, we concatenated $h$ linear projection matrices of BERT to one matrix to initialize the corresponding projection matrix in FPOA \& FPOA. We used the Adam optimizer \citep{kingma2014adam} with its default setting for optimization, with a batch size of 10. Dropout with a ratio of 0.3 was applied to the representations of BERT before projection. In Section \ref{section:Few-Shot Parsing}, we introduce two kinds of loss function, the negative likelihood loss and the margin loss. We found that the FPIO performed better with the negative likelihood loss, while the FPOA performed better with margin loss. The relationship between the loss function and the parsing performance will be analyzed in Section \ref{Ablation-study}. 

We randomly sampled 5 subsets with different numbers of annotated trees (10, 20, 40, 50, 80) from the PTB validation set\footnote{We trained FPOA and FPIO on validation set rather than the training set, for a fair comparison with \citet{shi2020role}} and removed the punctuation to retrain the two linear projection matrices of FPOA and FPIO as described in Section \ref{section:Few-Shot Parsing}. We compare our proposed FPOA and FPIO against our unsupervised method UPIO, a few-shot parsing method FSS \citep{shi2020role} and the Berkeley parser \citep{kitaev-etal-2019-multilingual}, which is also built on the base BERT model. Following \citet{shi2020role}, we only tuned the hyper-parameters of UPIO, FPOA and FPIO on 5 annotated trees of the validation set of PTB.\footnote{The layer on which FPOA and FPIO trained on was also selected according to 5 annotated trees from the validation set of PTB.}

 The results are shown in Table \ref{tab:few-shot parsing}. Although the Berkeley parser achieves high accuracy on supervised parsing, it is not comparable to FSS and our FPIO on few-shot parsing, especially when the number of training instances is $<=$ 20. Trained with only 10 annotated trees, FPIO outperforms FSS trained on the same number of  annotated trees by 4 $\text{F1}_{max}$ points. With 20 annotated trees, FPIO is better than our unsupervised UPIO by 11 $\text{F1}_{max}$ points and comparable to FSS trained on 50 annotated trees. Trained with 80 samples, FPIO achieves an improvement of 20 $\text{F1}_{\mu}$ points over UPIO. All these strongly suggest that we can train just the linear projection matrices of BERT (freezing all other parameters) with only a few annotated trees to obtain good parsing results. With the prior knowledge on syntax learned by the pretrained language models, our few-shot parsing framework (FPIO) can significantly reduce annotating overheads in building a syntactic corpus.

We observe that FPIO outperforms FPOA substantially on every group of training data. It is reasonable since FPIO exploits more information to estimate the likelihood of a tree, which may be noisy in the unsupervised setting, but makes the training easier in the few-shot setting.

\begin{table}

\centering
\caption{\label{tab:few-shot parsing} F1 scores of FPOA and FPIO trained on different numbers of annotated trees from the validation set of PTB, evaluated on the test data of PTB. FSS: a few-shot parsing method \citep{shi2020role}. Berkeley: a supervised parsing method \citep{kitaev-etal-2019-multilingual}}
\begin{tabular}{c c c c c c c c c c c}

\toprule
\multirow{2}{*}{Model} & \multicolumn{2}{c}{\textbf{10}} & \multicolumn{2}{c}{\textbf{20}} & \multicolumn{2}{c}{\textbf{40}} & \multicolumn{2}{c}{\textbf{50}} & \multicolumn{2}{c}{\textbf{80}} \\ 
\cmidrule(lr){2-3} \cmidrule(lr){4-5} \cmidrule(lr){6-7} \cmidrule(lr){8-9} \cmidrule(lr){10-11}
& $\text{F1}_{\mu}$ & $\text{F1}_{max}$ & $\text{F1}_{\mu}$ & $\text{F1}_{max}$ & $\text{F1}_{\mu}$ & $\text{F1}_{max}$ & $\text{F1}_{\mu}$ & $\text{F1}_{max}$ & $\text{F1}_{\mu}$ & $\text{F1}_{max}$ \\
\midrule
FPOA & 48.4 & 50.3 & 54.4 & 55.5 & 58.0 & 59.7 & 60.4 & 61.7 & 62.6 & 63.8 \\
FPIO & \textbf{53.7} & \textbf{57.4} & \textbf{60.4} & \textbf{61.6} & \textbf{65.3} & \textbf{66.6} & \textbf{67.0} & \textbf{68.4} & \textbf{68.8} & \textbf{69.4} \\
Berkeley & 27.2 & 32.8 & 38.3 & 40.3 & 47.8 & 49.9 & 54.0 & 55.4 & 57.6 & 59.7 \\
FSS & - & 53.4 & - & - & - & - & - & 61.2 & - & - \\
UPIO & - & 49.5 & - & 49.5 & - & 49.5 & - & 49.5 & - & 49.5 \\
\bottomrule
\end{tabular}
\end{table}

\subsection{Cross-lingual Parsing on SPMRL}\label{section:Cross-lingual parsing Results}
Although the unsupervised parsers in Section \ref{section:Unsupervised parsing results on PTB} were not trained on annotated trees, the hyper-parameters of them were tuned on the annotated trees in the validation set. \citet{shi2020role} show that the performance of most unsupervised parser, e.g., C-PCFG, heavily relies on tuning hyper-parameter on annotated trees. However, it is very expensive to obtain human-annotated trees for low-resource languages. If we do not have any annotated trees for these languages, one way is to resort to cross-lingual parsing. Following \citet{kimchart}, we first tuned the hyper-parameters of UPOA and UPIO on PTB, and then evaluated them on SPMRL dataset \citep{spmrl}. The SPMRL dataset contains 8 languages: Korean, German, Polish, Hungarian, Basque, French, Hebrew and Swedish.\footnote{The original SPMRL dataset contains 9 languages. Unfortunately, we don't have the license for Arabic treebank in this dataset. \citet{kimchart} have not used Arabic for cross-lingual experiment either.}

We evaluate both our unsupervised parser (UPOA, UPIO) and few shot parser (FPOA, FPIO) in the cross-lingual experiments. The hyper-parameter tuning for UPOA and UPIO is searching the best heads of the BERT for parsing on PTB. We selected all heads whose F1 score on the validation set is higher than 30, which were integrated for parsing on SPMRL by averaging the attention weight matrices. While the FPOA and FPIO were trained on the entire training set of PTB.

We compared our models with a multi-lingual unsupervised parser \citep{kimchart}, and the multi-lingual Berkeley parser. Both our models and compared baselines are based on the multi-lingual base BERT. For a fair comparison, all models are only trained/tuned on PTB. Raw texts of the training set of SPMRL were not used. Unlike all experiments on PTB, we did not remove punctuation on SPMRL. For the comparison with \citep{kimchart}, we report sentence-level F1 scores, while for the comparison with the Berkeley parser, we report corpus-level F1 scores with evalb.\footnote{As suggested by \citet{li-etal-2020-empirical}, we evaluate our model and Berkeley parser with evalb. However, \citet{kimchart} only report sentence-level F1 scores, hence we also report sentence-level F1 scores, for a fair comparison.}

The results are shown in Table \ref{tab:cross-lingual-on-spmrl}. It can be seen that both UPOA and UPIO outperform \citet{kimchart} substantially on 5 languages (``German'', ``Polish'', ``Hungarian'', ``Hebrew'', ``Swedish'') in terms of the sentence-level F1 score. Both our unsupervised methods (UPOA, UPIO) and few-shot (FPOA, FPIO) methods tuned/trained on PTB perform poorly on ``Basque'', which may indicate that the syntax of Basque is very different from that of English. The multi-lingual Berkeley parser trained on PTB performs poorly not only on Basque, but also on ``Korean'' and ``Polish''. Both our our unsupervised methods (UPOA, UPIO) and few-shot methods (FPOA, FPIO) significantly outperforms the Berkeley parser on 5 languages (``Korean'', ``German'', ``Polish'', ``Basque'',``Hebrew''). It can be seen that both our few-shot methods and the Berkeley parser perform well on ``French'' and ``Swedish'', which suggests that these two languages are similar to English in terms of syntax. We also observe that the performance gap between FPOA and FPIO is very small, which may be due to the fact that both of them are trained on the entire training set of PTB. In Section \ref{section:Few-shot Parsing on Different data size}, we will show that the performance gap between FPOA and FPIO decreases as the training data increases.
\begin{table}

    \centering
    \caption{\label{table: Cross-lingual parsing Results} F1 scores of models (tuned/trained on PTB) evaluated on 8 languages of SPMRL. The hyper-parameters of \citet{kimchart}, UPOA and UPOA were tuned on the validation set of PTB, while the parameters of FPOA, FPIO and Berkeley parser were trained on 80 trees from the training set of PTB. The results of \citet{kimchart} are produced with their top-down parsing algorithm.}
    \begin{tabular}{c c c c c c c c c}
    \toprule
        \textbf{Model} &  \textbf{Korean} & \textbf{German} & \textbf{Polish} & \textbf{Hungarian} & \textbf{Basque} & \textbf{French} & \textbf{Hebrew} & \textbf{Swedish}  \\
    \midrule
         & \multicolumn{8}{c}{Sentence-level F1} \\ \cmidrule(lr){2-9}
        \citet{Kim2020Are} & \textbf{45.7} & 39.3 & 42.3 & 38.0 & \textbf{41.1} & \textbf{45.5} & 42.8 & 38.7 \\
        UPOA & 43.6 & \textbf{40.9} & 46.4 & \textbf{38.5} & 27.2 & 40.2 & \textbf{44.5} & \textbf{42.02} \\
        UPIO & 40.3 & 40.0 &\textbf{46.8} & 35.5 & 24.7 & 38.8 & 42.8 & 41.9  \\
        \midrule
        & \multicolumn{8}{c}{Corpus-level F1} \\ \cmidrule(lr){2-9}
        UPOA     & \textbf{48.9}  & 41.59 & 49.71 & 42.3  & 36.43 & 38.63 & 45.08 & 42.97 \\
        UPIO     & 45.45 & 40.43 & 49.81 & 39.5  & 33.89 & 36.63 & 43.13 & 42.62 \\
        FPOA     & 46.4 & \textbf{42.2} & 60.2 & 43.7 & 29.5 & 50.5 & 60.1 & 57.8 \\
        FPIO     & 38.6  & 41.7 & \textbf{60.3} & 45.31 & \textbf{30.9}  & 50.9 & \textbf{60.6} & 57.8 \\
        Berkeley & 29.4 & 40.0 & 35.6 & \textbf{50.3} & 29.4 & \textbf{57.0} & 42.2 & \textbf{68.6} \\
        LB & 25.6 & 16.5 & 25.4 & 18.3 & 29.2 & 9.0 & 12.1 & 15.7 \\
        RB & 25.4 & 18.7 & 47.6 & 19.3 & 27.0 & 26.5 & 32.1 & 37.0 \\
    \bottomrule
    \end{tabular}
    \label{tab:cross-lingual-on-spmrl}
\end{table}

\section{Analysis}
\subsection{UPIO and UPOA tuned with Different Number of Annotated Trees}\label{section:unsupervised parsing with different data size}
In our unsupervised parsing experiments (Table \ref{parsing-on-subsets-of-wsj}), we selected heads for parsing using all annotated trees of the validation set. In this section, we follow the protocol proposed by \citet{shi2020role} to tune UPOA and UPIO with just a few annotated trees and investigate the impact of the number of annotated trees on the performance of UPIO and UPOA. Given a few annotated trees as the validation set, we first evaluate the F1 score of UPIO and UPOA on every head of BERT, and then average the attention matrices of the best 10 heads. Finally we evaluate the UPIO and UPOA with the averaged attention matrix on the test set. The results are shown in Figure \ref{fig:f1_of_up_diff_data_size}. It can be seen that even with only 1 annotated tree, our UPIO and UPOT obtain an F1 score higher than 49, and that increasing the size of validation data does not improve F1 score significantly, which indicates that our UPIO and UPOA can be tuned well with very few annotated trees.

\subsection{FPOA and FPIO Trained with Different Numbers of Annotated Trees}\label{section:Few-shot Parsing on Different data size}
In the few-shot parsing experiments of Section \ref{section:Few-shot parsing results on PTB}, we trained FPOA and FPIO with annotated trees of size (10, 20, 40, 80). In this section, we trained FPOA and FPIO with more annotated trees, and visualize the results in Figure \ref{fig:f1_of_diff_data_size}. It can be seen that the F1 score grows fastest when the number of annotated trees is < 1000. As the size of training data increases, the growth of F1 score slows down. We also observe that FPIO outperforms the FPOA all the time. However, the gap between the performance of the two model decrease as the training data size increases. FPIO exploits more information from the self-attention weight matrix for parsing, which may explain why FPIO performs better than FPOA training on few examples. However, training with enough data, the attention matrix in FPOA can well approximate the distance matrix of its corresponding tree. Therefore, FPOA can perform as well as the FPIO with enough training data.

\subsection{Few-shot Parsing on Different Layers}\label{section:Few-shot Parsing on Different Layers}
\begin{figure}
\centering
    \begin{subfigure}[b]{0.4\linewidth}	
        \centering
        \includegraphics[width=1.0\linewidth]{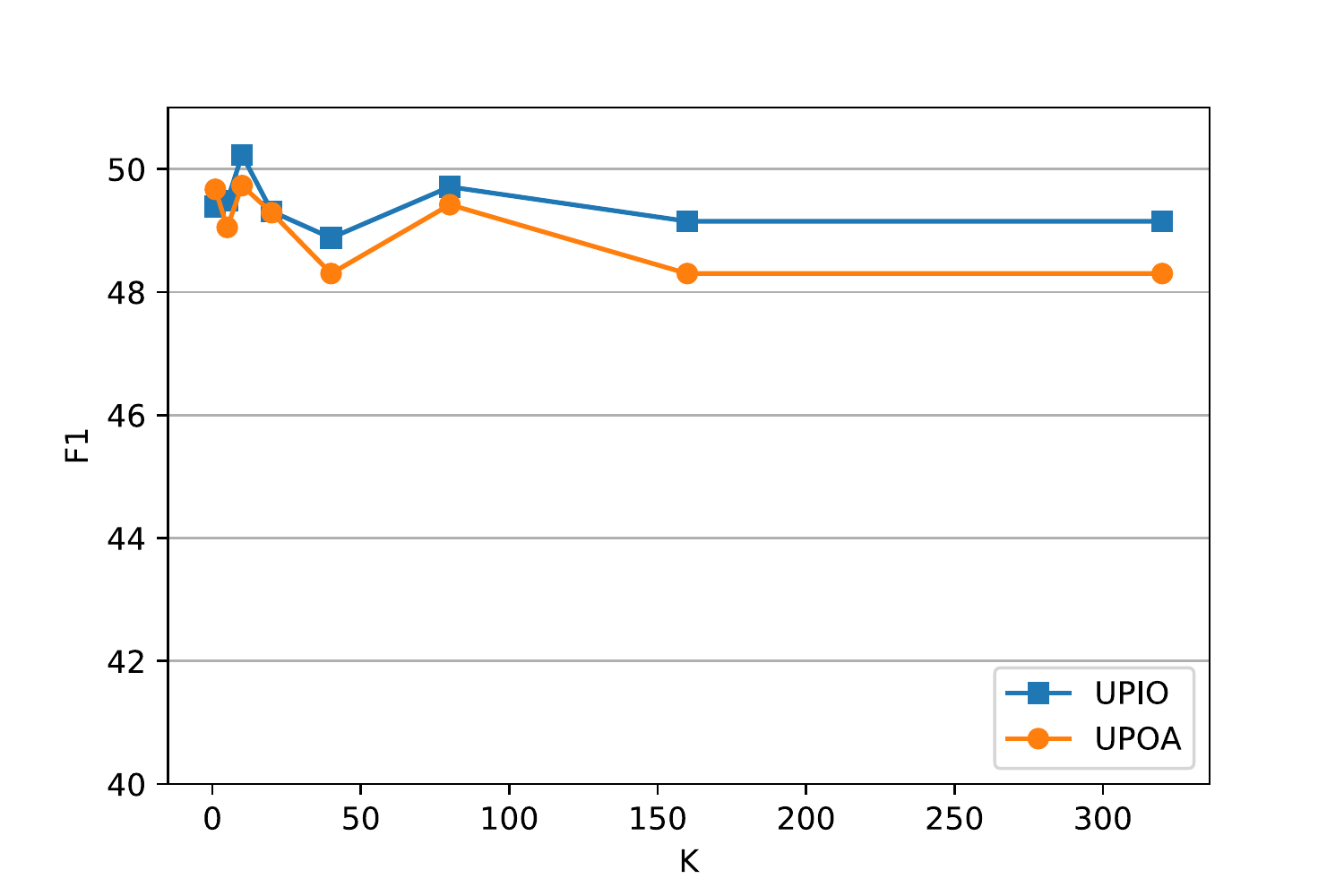} 
        \caption{\label{fig:f1_of_up_diff_data_size}}
    \end{subfigure}
    \begin{subfigure}[b]{0.4\linewidth}	
        \centering
        \includegraphics[width=1.0\linewidth]{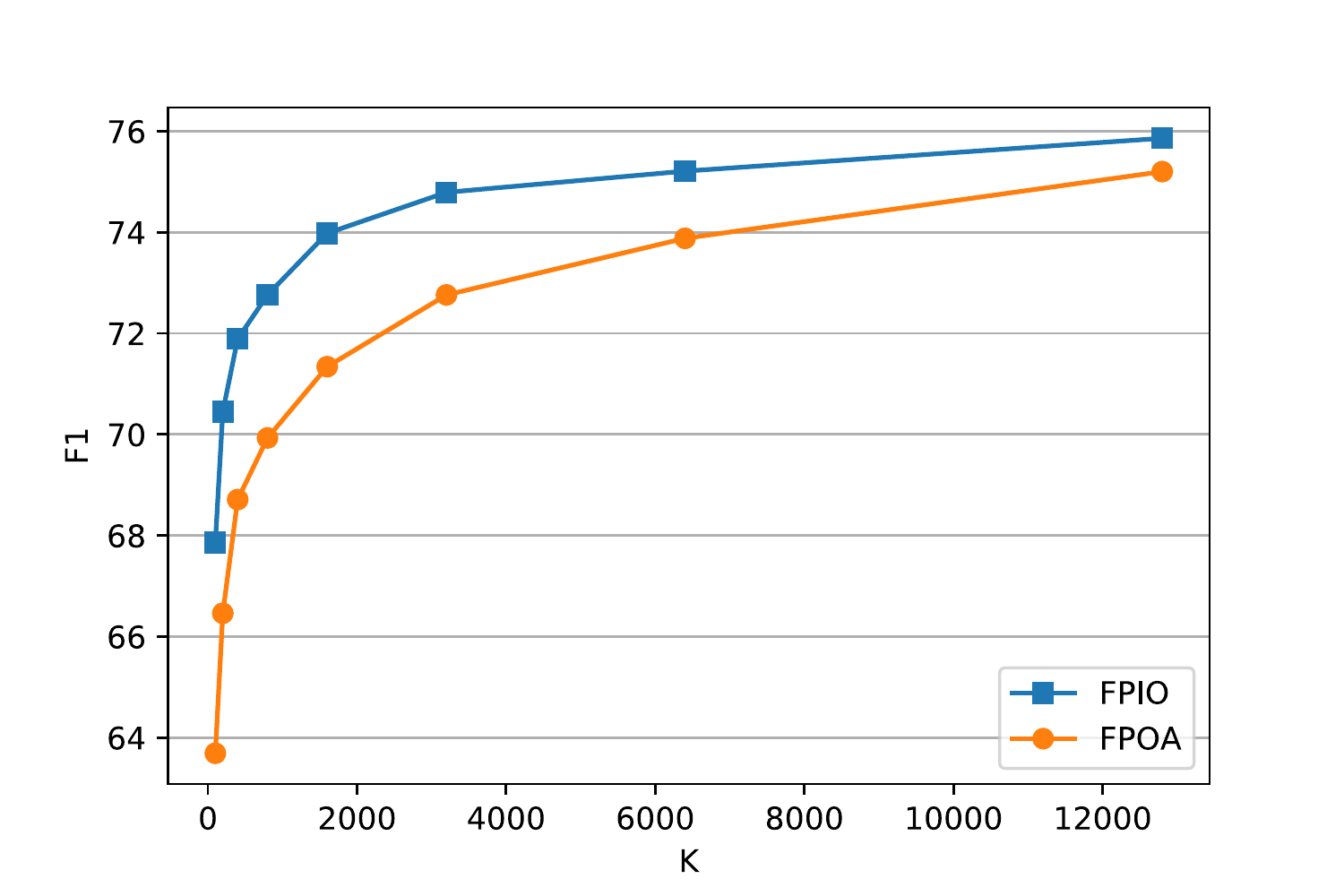} 
        \caption{\label{fig:f1_of_diff_data_size}}
    \end{subfigure}
    \begin{subfigure}[b]{0.4\linewidth}	
        \centering
        \includegraphics[width=1.0\linewidth]{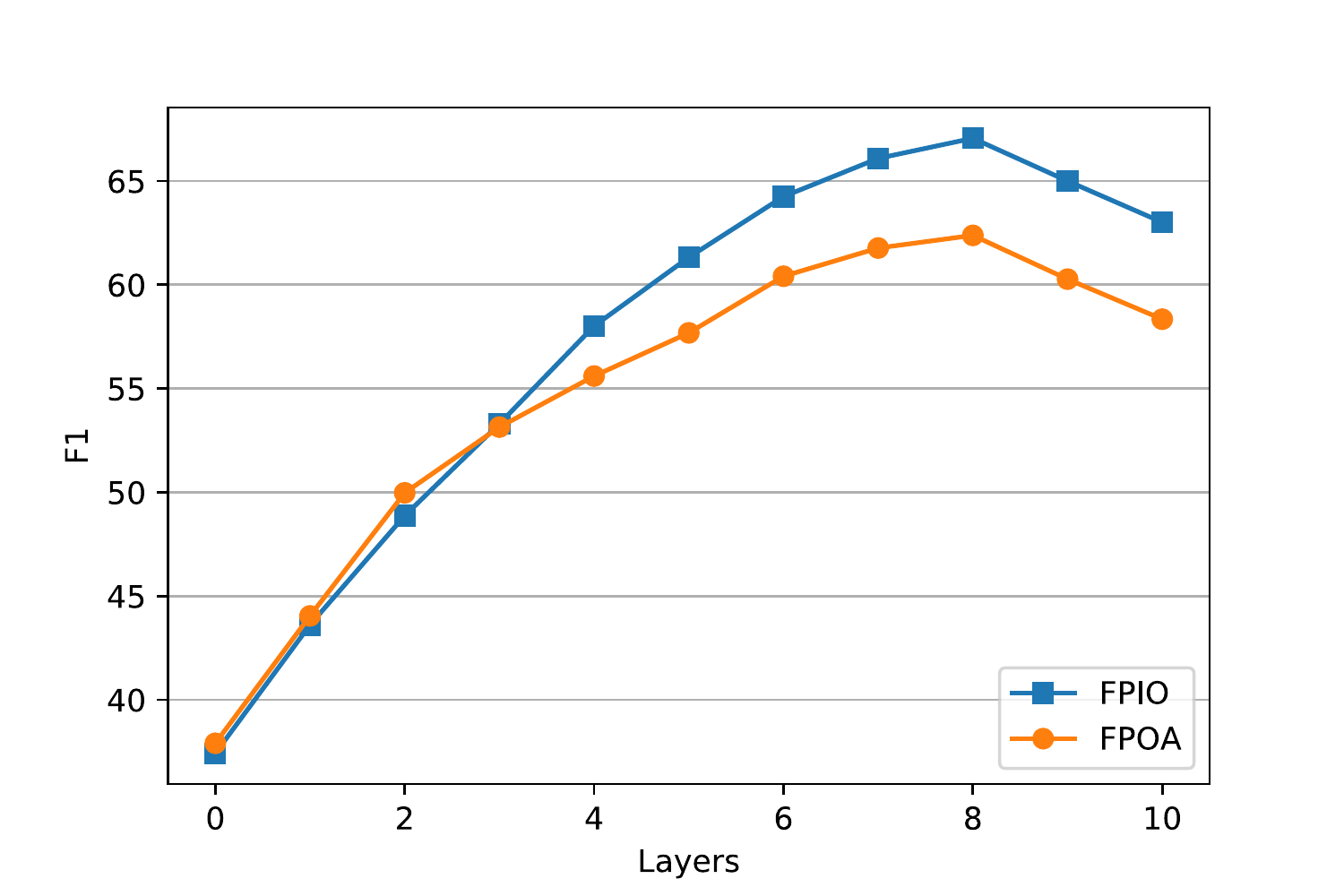} 
        \caption{\label{fig:f1_of_diff_layer}}
    \end{subfigure}
    \begin{subfigure}[b]{0.4\linewidth}
        \centering
        \includegraphics[width=1.0\linewidth]{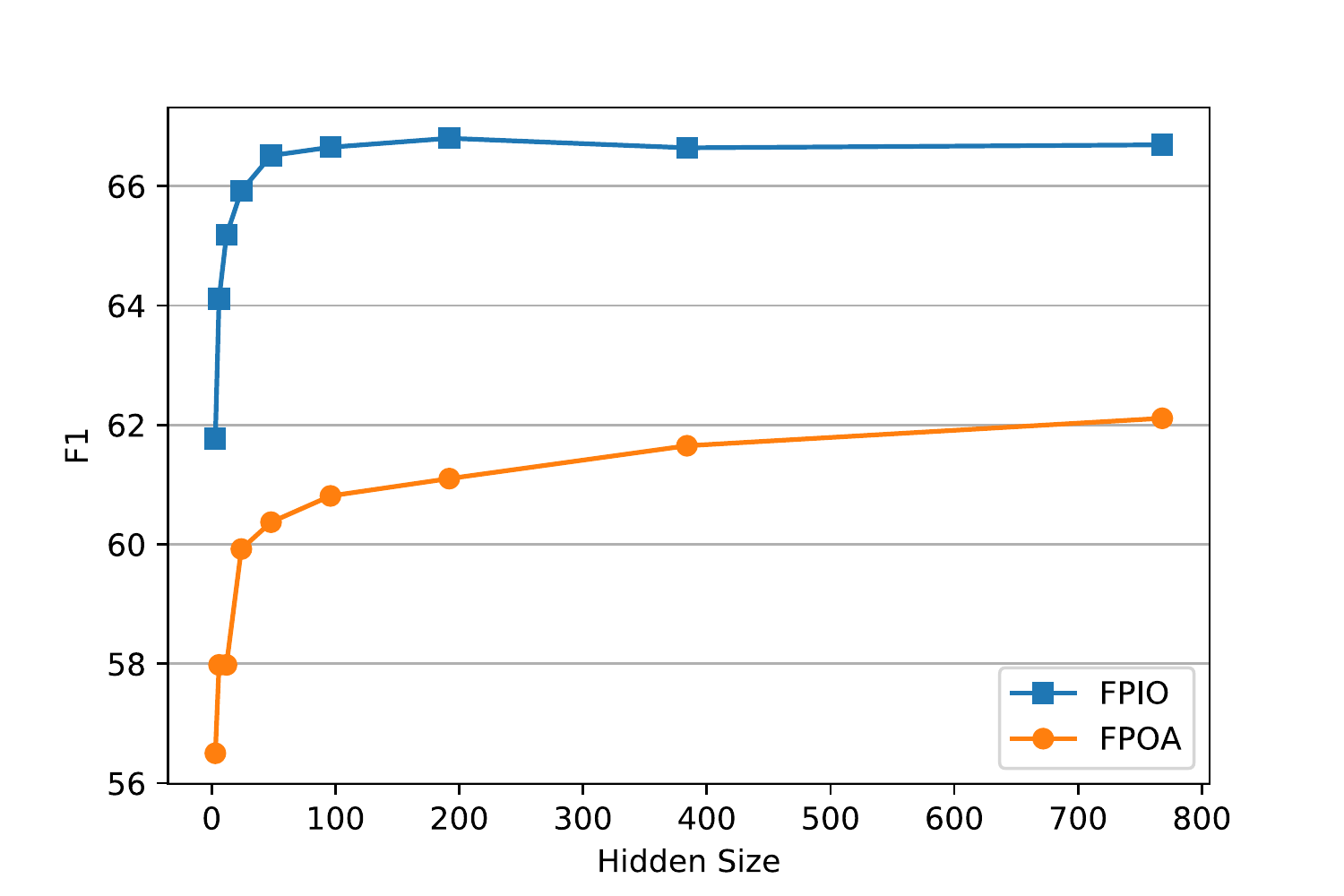}
        \caption{\label{fig:f1_of_diff_hid_size}}
    \end{subfigure}
\caption{(a): The F1 scores of UPOA and UPIO tuned on different numbers of annotated trees (1, 5, 10, 20, 40, 80, 160, 320). (b): The learning curves of FPOA and FPIO trained on annotated trees of different sizes (100, 200, 400, 800, 1600, 3200, 6400, 12800). (c): The F1 scores of FPOA and FPIO on different layers of BERT. (d): The F1 scores of FPOA and FPIO with linear projections of different size. All F1 scores are evaluated on the test set of PTB. The FPOA and FPIO in Figure (c) and (d) are trained with 80 annotated trees from training set of PTB.}
\end{figure}

We compared the performance of FPOA and FPIO with different layers of BERT trained on PTB. The results are shown in Figure \ref{fig:f1_of_diff_layer}. The performance curve of FPOA and FPIO varying with different layers resonates with the finding by \citet{hewitt-manning-2019-structural} who train a probe on BERT embeddings to produce dependency trees. Previous works \citep{tenney-etal-2019-bert,jawahar-etal-2019-bert} have also found that the intermediate layers of BERT perform better for syntax tasks than other layers. At low layers (i.e., layer 0, 1, 2, 3), the performance of FPOA is comparable with (even better than) the performance of FPIO. However, on high layers, especially the 8th layer, the FPIO outperforms FPOA significantly. Since both FPOA and FPIO have the best performance at the 8th layer, the FPOA and FPIO in Table \ref{tab:few-shot parsing}, \ref{table: Cross-lingual parsing Results} and Figure \ref{fig:f1_of_diff_data_size}, \ref{fig:f1_of_diff_hid_size} are all trained on the 8th layer.

\subsection{Few-Shot Model Training with Different Size of Projection Dimension}
The linear projection in the multi-head self-attention mechanism projects BERT hidden representations onto a new sub-space. In our experiments in Section \ref{section:Few-shot parsing results on PTB} \& \ref{section:Few-shot Parsing on Different Layers}, the dimension of projected representations in FPOA and FPIO is $d_{\text{model}}$. In this section, we investigated the impact of different projection dimension on the performance of parsing. The results are shown in Figure \ref{fig:f1_of_diff_hid_size}. It seems that increasing the projection dimension beyond 100 leads to small gains in parsing performance, especially for FPIO. This suggests that the syntactic information learned from a few annotated trees can be encoded in a low dimension space. 

\subsection{Contribution of Different Heads to few-shot parsing}
\begin{figure}
    \centering
    \includegraphics[width=0.3\textwidth]{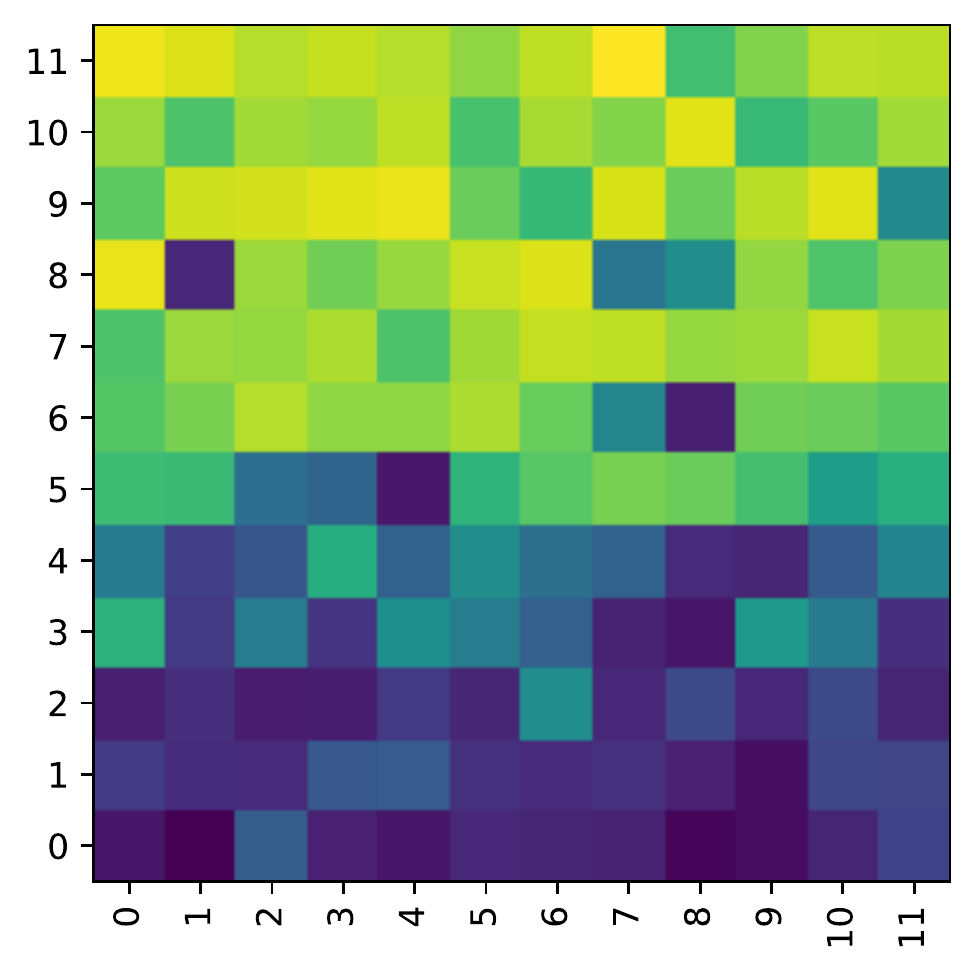}
    \caption{\label{fig:contribution-of-heads}Linear combination weights of different heads in FPIO. The pixel in the $i$th row, and $j$th column indicates the weight of the $j$th head of $i$th layer. The darker the color, the lower the weight.}
\end{figure}
In Section \ref{section:Unsupervised parsing results on PTB}, we averaged 3 attention heads which are significantly better than other heads in unsupervised experiments. But the attention matrices of these three heads may be similar to each other, and other heads may also provide useful information for parsing. Therefore, in this Section, we exploit all attention metrics in BERT by weighted linear combination for parsing:
\begin{equation}
    \mathbf{A}=\sum_i w_i \mathbf{A}_{h_i}
\end{equation}
$w_i$ is the weight for the $i$th head, while $\mathbf{A}_{h_i}$ is the attention matrix of the $i$th head. $\mathbf{A}$ can be used by our unsupervised/few-shot methods for unsupervised/few-shot parsing. Since it is difficult to obtain $w_i$ without annotated trees, we initialize $w_i$ with the uniform distribution, and update them in FPIO through backward propagation. We display the weights learned in FPIO in Figure \ref{fig:contribution-of-heads}. It is shown that the weights of heads in higher layers (7,8,9,10,11,12) are larger than those in low layers (1,2,3,4,5). We could also reach the same conclusion from Figure \ref{fig:parse-on-diff-heads} in Section \ref{section:Unsupervised parsing with different heads}, but Figure \ref{fig:contribution-of-heads} shows the contribution/importance of heads more clearly. We also observe that there exists some heads in high layers, e.g., the 1st head in the 9th layer, which have very small contribution. These heads may be irrelevant to syntax.

\subsection{Ablation Study}\label{section:Ablation study}
\begin{table}
    \centering
    \caption{\label{Ablation-study}Ablation study on (UPOA, UPIO) and (FPOA, FPIO). FPOA and FPIO was trained on 80 annotated trees from PTB training data. F1 scores were evaluated on the test data of PTB.}
    \begin{tabular*}{0.6\textwidth}{c @{\extracolsep{\fill}} ccc}
    \toprule 
    \textbf{Model} & \textbf{Factor} & \textbf{F1}\\ \midrule
    \multirow{4}{*}{\textbf{UPOA}} & greedy parsing & \textbf{50.9} \\
    & chart parsing & 22.4\\
    & No ensemble & 48.4 \\
    & No fine-tuning & 49.6 \\
    \midrule
    \multirow{5}{*}{\textbf{UPIO}} & greedy parsing & 45.5\\
    & chart parsing & \textbf{50.3}\\
    & No ensemble & 46.7 \\
    & No fine-tuning & 48.8\\
    & No outside association & 41.1 \\
    \midrule
    \multirow{2}{*}{\textbf{FPIO}} & Mle loss & \textbf{67.0} \\
    & Margin loss & 60.2 \\
    \midrule
    \multirow{2}{*}{\textbf{FPOA}} & Mle loss & 57.3 \\
    & Margin loss & \textbf{61.7} \\
    \bottomrule
    \end{tabular*}
\end{table}
We conducted ablation study on (UPOA, UPIO) and (FPOA, FPIO) to better understand the impact of several techniques used in these models. The results are displayed in Table \ref{Ablation-study}. ``greedy parsing'' and ``chart parsing'' are the parsing algorithms used in UPIO and UPOA, which have been introduced in Section \ref{section:Parsing with Split Score}. ``ensemble'' for UPIO and UPOA indicates that we integrate the best three attention heads for parsing as discussed in Section \ref{section:Unsupervised parsing with different heads}. ``Fine-tuning'' for UPIO and UPOA denotes that we fine-tune BERT on raw texts of PTB training set. ``outside association'' for UPIO denotes the score calculated in Eq. (\ref{inside-outside-rel-score}). ``Mle loss'' for FPIO and FPOA denotes the negative likelihood loss formulated in Eq. (\ref{eq:mle-loss}), while ``Margin loss'' is formulated in Eq. (\ref{eq:margin-loss}).

From Table \ref{Ablation-study}, we can observe that UPOA with the chart parsing algorithm can not produce valid trees. The reason is that the outside association of sub-spans can not be used for the split decision of the current span. However, UPIO achieves its best performance with the chart parsing. The key difference between UPIO and UPOA is that UPIO exploits the inside association. Therefore UPIO with chart parsing is better than that with greedy parsing because of the inside association. Unlike the outside association, the inside association of sub-spans is relevant to that of the current span.

The ensemble of different attention heads has a great impact on UPOA and UPIO as it combines different views from different attention heads for parsing. Fine-tuning BERT on raw texts also helps UPOA and UPIO. As mentioned before, \citet{marecek-rosa-2018-extracting} also exploit the self-attention weight matrix for unsupervised parsing, but they haven't considered the outside association score. In our ablation study, we removed the outside association score to simulate their method. The performance of UPIO significantly drops by 8.2 F1, which indicates that the outside association score is a very important indicator for a span to be a constituent. 

It is interesting that FPIO performs better with the negative likelihood loss (MLE), while the FPOA performs better with the margin loss. The negative likelihood loss forces the probability of the ground truth to be as high as possible, while the margin loss converges as long as the margin constraint is satisfied. Therefore, the model is easier to overfit the training data with the negative likelihood loss than with the margin loss. FPOA only exploits the outside association from the attention weight matrix, which makes it easier for FPOA to overfit the few training data. This may explain why FPOA performs better with the margin loss. By contrast, FPIO exploits more information (inside association) from the attention weight matrix than FPOA, which makes it harder to overfit the few training data, even with the negative likelihood  loss.

\subsection{Visualization of Attention Weight Matrix}\label{section:Visualization of Attention Weight Matrix}
\begin{figure}[t]
\centering
\begin{subfigure}[b]{0.4\linewidth}	
\centering
\includegraphics[width=1.0\linewidth]{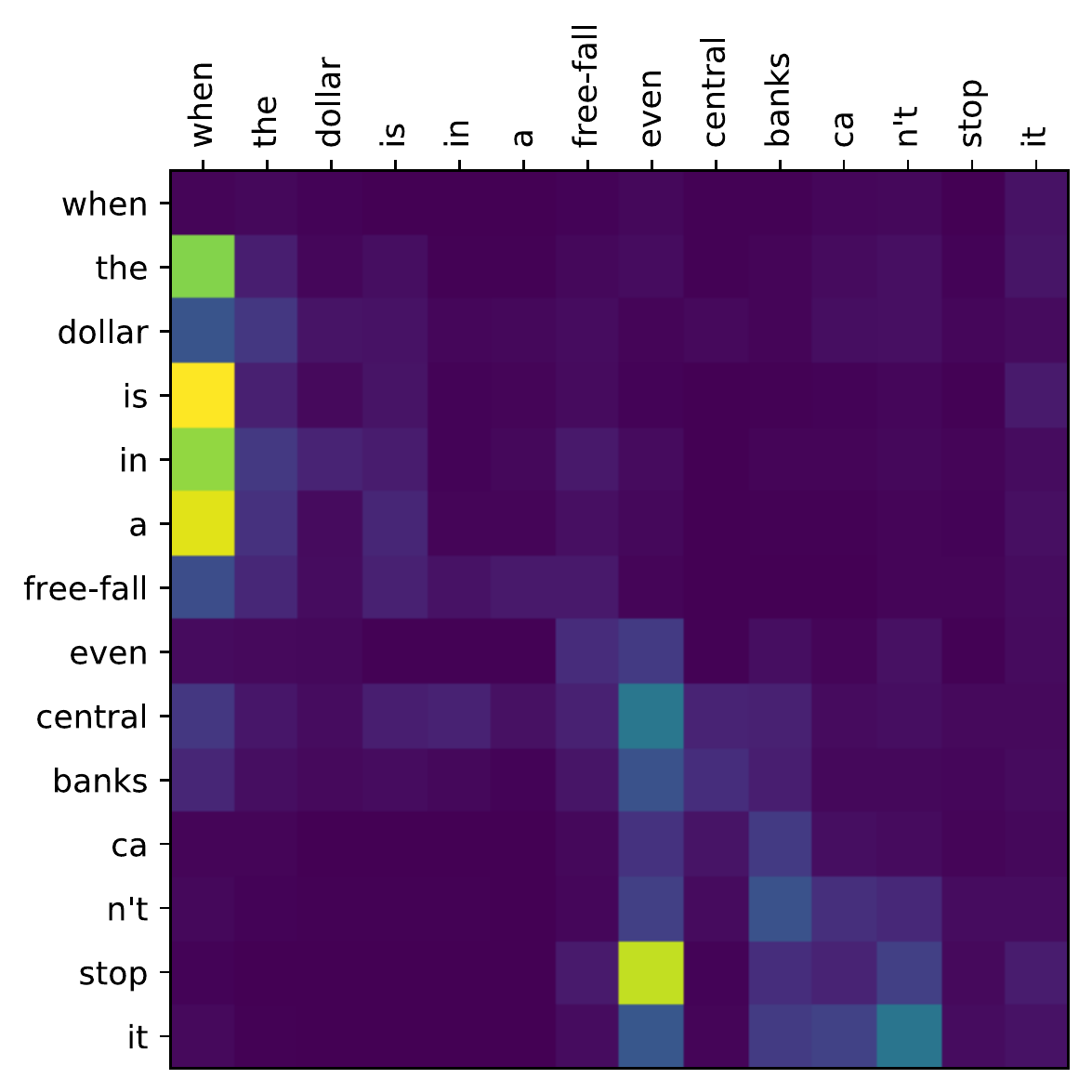} 
\caption{}
\label{fig:subim1}
\end{subfigure}
\begin{subfigure}[b]{0.4\linewidth}
\centering
\includegraphics[width=1.0\linewidth]{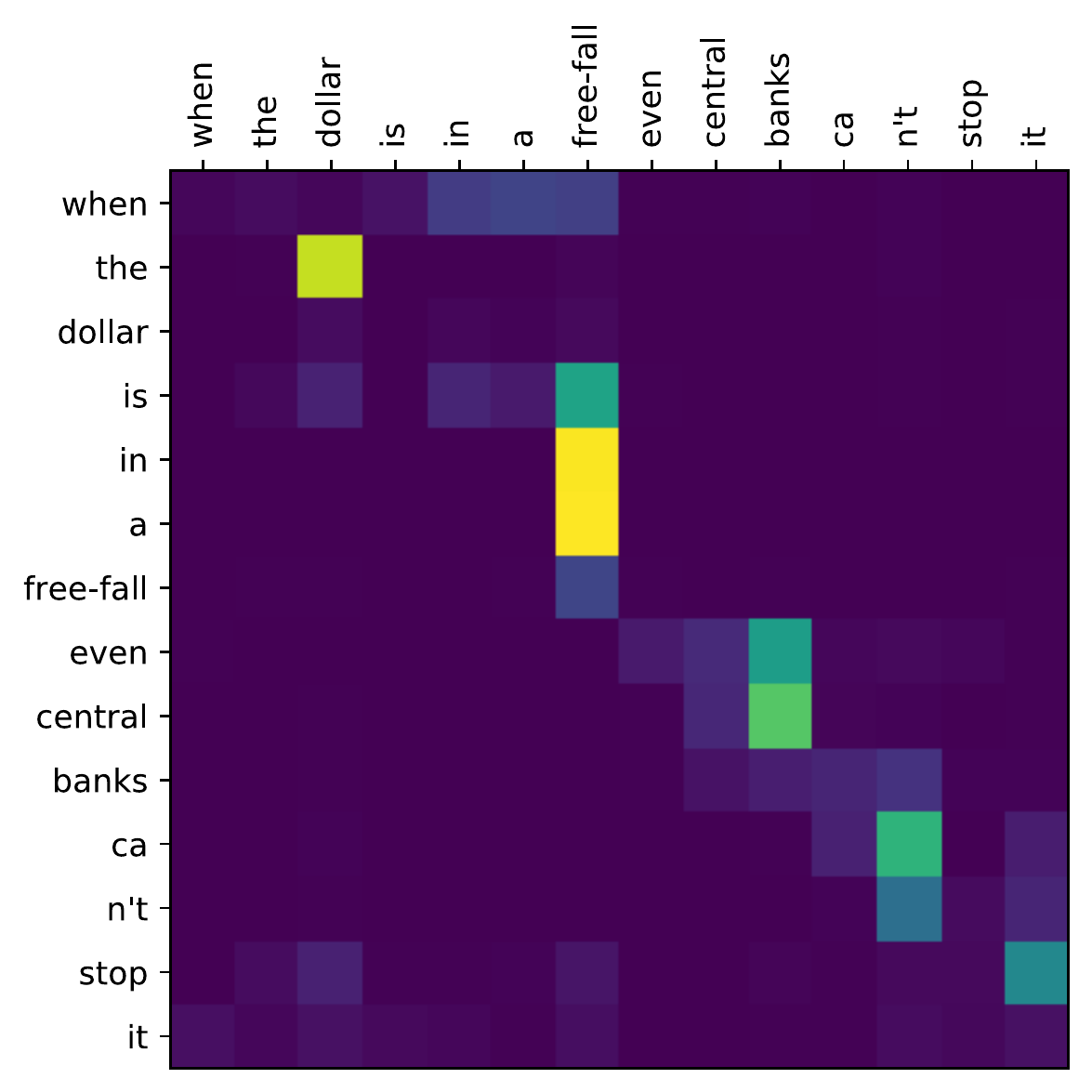}
\caption{}
\label{fig:subim2}
\end{subfigure}
\centering
\begin{subfigure}[b]{0.4\linewidth}
\includegraphics[width=1.0\linewidth]{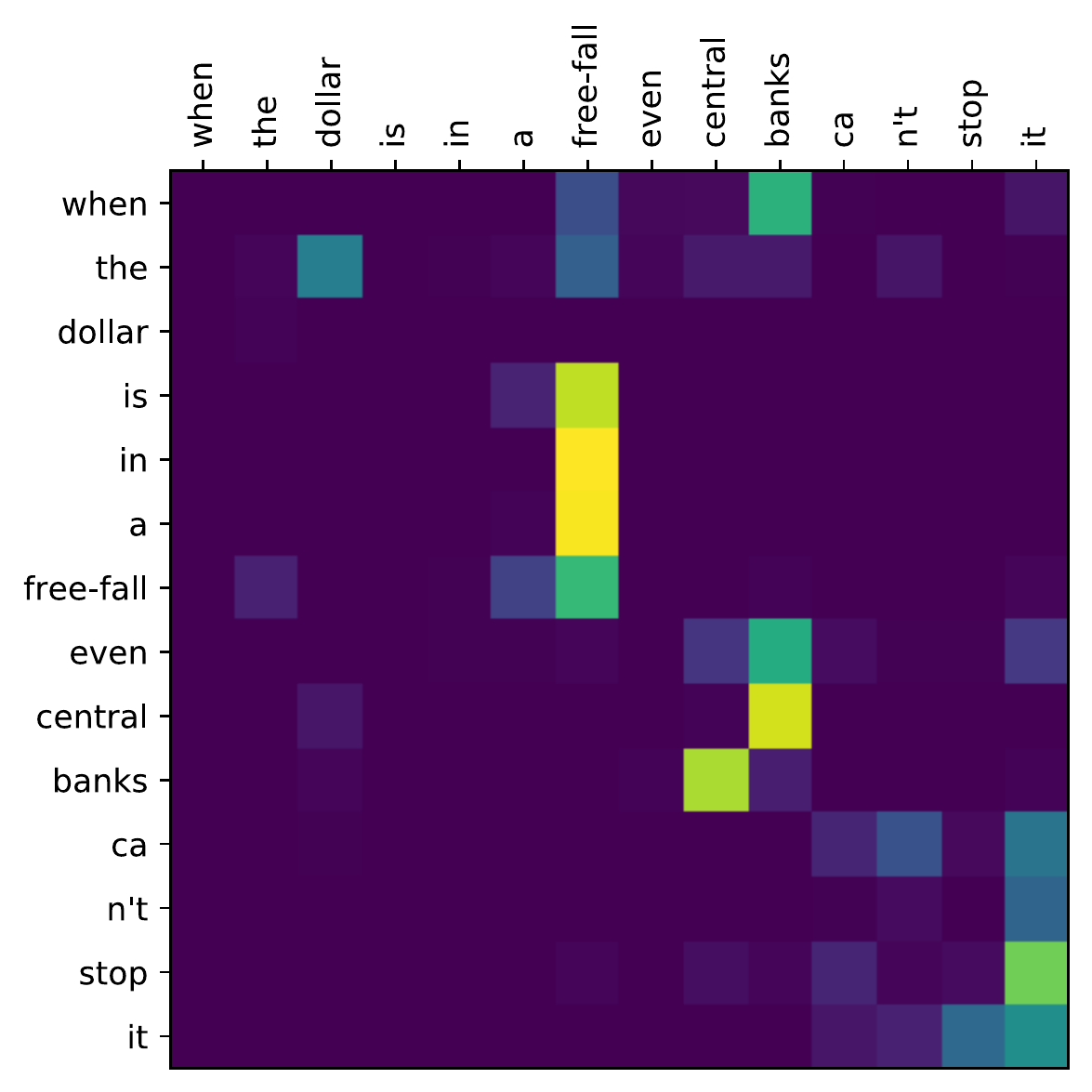}
\caption{}
\label{fig:subim3}
\end{subfigure}
\begin{subfigure}[b]{0.4\linewidth}
\centering
\includegraphics[width=1.0\linewidth]{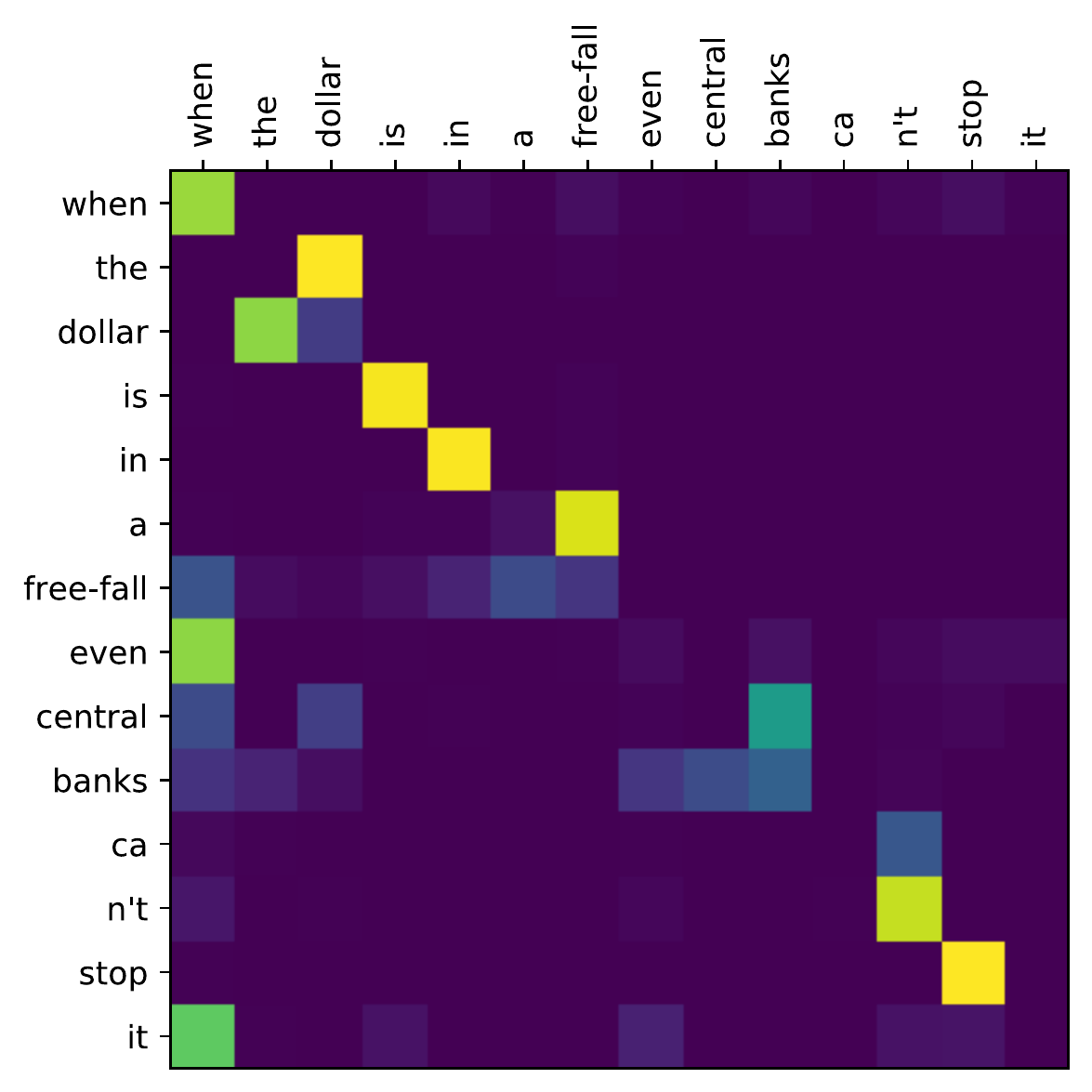}
\caption{}
\label{fig:subim4}
\end{subfigure}
\caption{Visualization of attention weight matrix. (a) is from a randomly selected head of BERT. (b) is from the 10th head of the 7th layer which is the best attention head for UPIO and UPOA on PTB. (c) is from the FPOA trained on 80 annotated trees of the PTB training set. (d) is from the FPIO trained on 80 annotated trees of the PTB training set.}
\label{fig:visual-attention}
\end{figure}

Our models are built upon the attention weight matrix of BERT. It is therefore interesting to visualize and compare the attention weight matrices from different parsing models for the same sentence. We show an example in Figure \ref{fig:visual-attention}, which visualizes four different attention weight matrices from the UPOA, UPIO, FPOA and FPIO. The attention weight matrix (\ref{fig:subim1}) is from a randomly chosen head, from which we can observe that the sentence is split into two sub-spans at the position of word ``even''. The attention weight matrix (\ref{fig:subim2}) is from the attention head that achieves the best performance in the unsupervised parsing model UPOA and UPIO, from which we can observe that the attention weights inside some constituent, e.g., ``in a freeball'', are significantly higher than that outside the constituent. 

The attention weight matrix (\ref{fig:subim3}) from FPOA is very similar to the attention weight matrix (\ref{fig:subim2}) from unsupervised models. However, the attention weight matrix (\ref{fig:subim4}) from FPIO is very different from the attention matrices from FPOA (\ref{fig:subim3}) and unsupervised models (\ref{fig:subim2}). We also observe that the attention weights from FPIO (\ref{fig:subim3}) focus on the diagonal, from which we can also recognize some constituents easily, e.g., ``the dollar'', ``a free-fall''. These suggest that our parsing models are easily interpretable as constituents can be detected from the used/retrained attention weight matrices.

\section{Conclusions}
In this paper, we have presented two unsupervised constituent parsing methods UPOA \& UPIO based on the self-attention weight matrix of the pretrained language model BERT. UPOA splits a span according to the outside association, while the UPIO splits a span according to the inside association together with outside association. The proposed UPOA is further extended to FPOA while the UPIO to FPIO to improve the attention weight matrix by retraining the linear projection matrices of the query and key with only a few annotated trees. 
Our unsupervised parser UPOA and UPIO achieve results comparable to the state-of-the-art results on short sentences (length <= 10) of PTB. Our few-shot parser FPIO trained with a few annotated trees outperforms the compared few-shot and supervised parser. Both our unsupervised parser and few-shot parser are better than previous methods on most languages in cross-lingual experiments. We have also found that UPOA that only exploits outside association is better than UPIO that exploits both inside and outside association on long sentences, while FPIO extended from UPIO performs better than FPOA extended from UPOA in few-shot experiments. Our unsupervised methods require very few annotated trees for hyper-parameter tuning. Tuned on only 1 annotated tree, UPOA and UPIO achieves substantially better results than other fully unsupervised parser. We have further analyzed the impact of the training data, training layer and projection dimension on FPOA and FPIO, and have found that 1) the gap between FPOA and FPIO narrows as the training data increases, 2) FPOA and FPIO trained on immediate layers perform better than that on the other layers, and 3) The syntactic information learned from a few annotated trees is encoded in a low dimension space. The visualization on the attention weight matrices makes our proposed parsers interpretable, as constituents can be detected from the attention weight matrices.

\section*{Acknowledgements}
The present research was supported by the National Key Research and Development Program of China (Grant No. 2019QY1802).

\bibliographystyle{unsrtnat}  
\bibliography{references}

\end{document}